\documentclass[]{article}
\usepackage[margin=2cm]{geometry}
\usepackage{url}
\usepackage{graphicx}
\usepackage[export]{adjustbox}
\usepackage[font=footnotesize,labelfont=bf]{caption}
\usepackage{mathtools}
\usepackage{bm}
\usepackage[T1]{fontenc}
\usepackage{amsmath}
\usepackage{amssymb}
\usepackage{hyperref}
\usepackage{fancyhdr}
\usepackage[dvipsnames]{xcolor}
\usepackage[utf8]{inputenc}
\usepackage{authblk}
\usepackage{rotating}
\usepackage{comment}
\usepackage{bbm}
\usepackage{soul}
\usepackage{subcaption}
\usepackage[normalem]{ulem}
\usepackage{multirow}
\usepackage{pdflscape}
\usepackage{hhline}
\usepackage{float}
\usepackage{booktabs}
\usepackage{lineno}
\usepackage{tabularx}
\usepackage{array}
\usepackage{amsmath}
\usepackage{xr-hyper}
\usepackage{hyperref}
\usepackage{xfrac}
\usepackage{makecell}
\usepackage{ragged2e}

\usepackage[sorting=none,
            style=numeric-comp,
            bibstyle=ieee,
            url=false,
            doi=false,
            isbn=false,
            language=english,
            clearlang=true,
            maxbibnames=20,
            arxiv=false,
            giveninits=true]{biblatex}

\usepackage{biblatex-shortfields}

\DeclareFieldFormat{ieeecase}{#1}
\DeclareNameAlias{sortname}{first-last}

\renewbibmacro{in:}{}

\DeclareFieldFormat[article]
{volume}{#1} 

\DeclareFieldFormat[article, inbook, incollection, inproceedings, misc, thesis, unpublished]
{number}{(#1)}

\DeclareFieldFormat[article,inproceedings]
{pages}{#1}
\DeclareFieldFormat[inbook, incollection]
{pages}{p. #1}

\renewbibmacro*{volume+number+eid}{
  \printfield{volume}\printfield{number}:
}

\AtEveryBibitem{\clearfield{month}}
\AtEveryBibitem{\clearfield{day}}
\AtEveryBibitem{%
  \clearlist{language}%
}

\bibliography{ref.bib}

\title{TimeSeries2Report prompting enables adaptive large language model management of lithium-ion batteries}
\date{\vspace{-7ex}}

\author[1]{Jiayang Yang}
\affil[1]{\textit{State Key Laboratory of Industrial Control Technology, College of Control Science and Engineering, Zhejiang University, Hangzhou, 310027,
 China}}
\author[2]{Martin Guay}
\author[2]{Zhixing Cao\footnote{z.cao@queensu.ca}}
\author[1]{Chunhui Zhao\footnote{chhzhao@zju.edu.cn}}
\affil[2]{\textit{Department of Chemical Engineering, Queen's University, Kingston, Ontario K7L 3N6, Canada}}

\begin{document}
\renewcommand{\figurename}{Fig.}
\renewcommand{\thefigure}{\arabic{figure}}
\captionsetup[figure]{labelformat=simple}

\maketitle

\begin{abstract}
Large language models (LLMs) offer promising capabilities for interpreting multivariate time-series data, yet their application to real-world battery energy storage system (BESS) operation and maintenance remains largely unexplored. Here, we present TimeSeries2Report (TS2R), a semantic translation framework that converts raw lithium-ion battery operational time-series into structured, semantically enriched reports, enabling LLMs to reason, predict, and make decisions in BESS management scenarios. TS2R encodes short-term temporal dynamics into natural language through a combination of segmentation, semantic abstraction, and rule-based interpretation, effectively bridging low-level sensor signals with high-level contextual insights. We benchmark TS2R across both lab-scale and real-world datasets, evaluating report quality and downstream task performance in anomaly detection, state-of-charge prediction, and charging/discharging management. Compared with vision-, embedding-, and text-based prompting baselines, report-based prompting via TS2R consistently improves LLM performance in terms of across accuracy, robustness, and explainability metrics. Notably, TS2R-integrated LLMs achieve expert-level decision quality and predictive consistency without retraining or architecture modification, establishing a practical path for adaptive, LLM-driven battery intelligence.

\end{abstract}

\section{Introduction}
From portable electronics \cite{liang2019review} to electric vehicles \cite{chen2019recycling,harper2019recycling,li2019data} and grid-scale renewable energy storage systems \cite{arteaga2017overview,jiang2025battery}, lithium-ion batteries (LIBs) power nearly every facet of modern life. As deployment scales from individual cells to gigawatt-hour battery energy storage systems (BESS), ensuring the safety, reliability, and efficiency of such large, heterogeneous networks has become increasingly complex and mission-critical. Effective battery management requires precise, real-time monitoring of cell states including state of charge (SOC) \cite{adaikkappan2022modeling,shen2017co}, state of health (SOH) \cite{hu2020battery,rhyu2025systematic,severson2019data}, and temperature \cite{cao2025model,mei2023operando} under highly dynamic load conditions, evolving degradation pathways, and environmental variability. Yet the coupled electrochemical, thermal, and mechanical dynamics across thousands of interacting cells make predictive maintenance and adaptive control fundamentally challenging \cite{mama2025comprehensive,che2022state,xie2022faults}.

Artificial intelligence (AI) has emerged as a powerful enabler of advanced battery management, supporting a wide range of tasks from diagnostics to optimization. Deep learning models such as long short-term memory (LSTM) networks \cite{tan2019transfer}, convolutional neural networks (CNNs) \cite{gu2023novel,mazzi2024lithium}, and physics-informed neural networks (PINNs) \cite{wang2024physics,nascimento2021hybrid} have been successfully applied to estimate SOH and forecast battery aging. Unsupervised methods including autoencoders \cite{cao2025model,zhang2023realistic}, Gaussian mixture models \cite{yang2025toward}, and isolation forests \cite{jiang2022fault} have proven effective for anomaly detection, identifying faults such as overcharging, short circuits, and connection failures by extracting latent patterns from operational data. Reinforcement learning approaches \cite{wang2025data,park2022deep,hao2023adaptive} have further enabled dynamic energy and charging strategy optimization under varying load conditions.

Despite these advances, most AI-based battery management systems remain task-specific, data-hungry, and difficult to generalize across chemistries, architectures, or usage scenarios. Their limited interpretability and retraining overhead hinder deployment in safety-critical applications where transparency and adaptability are essential. These limitations are particularly pronounced in real-world operation and maintenance (O$\&$M) settings, where regulatory compliance, reliability, and human oversight are non-negotiable. While explainable AI (XAI) techniques such as SHapley Additive exPlanations (SHAP) \cite{lee2022state,zhao2025lithium} and Layer-wise Relevance Propagation (LRP) \cite{wang2023explainability} have been proposed to enhance model interpretability, they typically yield post hoc numerical feature attributions that require significant cognitive effort to interpret – falling short of true human-AI collaboration.

Large language models (LLMs) \cite{naveed2025comprehensive,li2025self,zuo2025large,majumder2024exploring,achiam2023gpt,zhao2025align} offer a compelling alternative, enabling flexible reasoning, contextual understanding, and natural language generation across diverse tasks. Their general-purpose nature opens the door to a new paradigm in battery management – one centered on explainability, adaptability, and task generalization. However, mainstream LLMs such as ChatGPT \cite{achiam2023gpt} are designed primarily for text-based conversations and lack native support for interpreting continuous, multivariate, and temporally structured time-series data. As highlighted in recent studies \cite{jiang2024empowering,wang2025itformer,11268252}, this architectural mismatch limits their utility in time-series-dominant domains such as battery O$\&$M, where key tasks like SOH estimation and fault diagnosis depend on rich temporal dynamics. Recent developments in multimodal LLMs, such as ChatTS \cite{xie2024chatts} and Time-LLM \cite{jin2023time}, address this gap by learning shared embedding spaces for time-series and text, allowing LLMs to treat signal segments as pseudo-tokens. However, these models require extensive retraining and deep natural language processing (NLP) expertise, which limits their accessibility and practical use in the battery domain.

In this work, we introduce TimeSeries2Report (TS2R), an automated signal-to-semantics-to-report framework that enables LLMs to perform explainable battery management without specialized architecture or retraining. Our method converts raw operational time-series data such as voltage, current, temperature, and capacity into structured, human-readable reports that serve as textual proxies for the original signals. These reports are constructed using a universal template that generalizes across multivariate and multi-cell LIB systems, and can be directly integrated into off-the-shelf LLMs via prompt engineering. By translating quantitative dynamics into semantically rich descriptions, TimeSeries2Report empowers LLMs to carry out key O$\&$M tasks including condition assessment, anomaly detection, and decision support with strong interpretability and minimal infrastructure cost. Notably, our approach requires no model training, making it lightweight, scalable, and accessible to users beyond the AI community. To support reproducibility and further research, we release the first large-scale paired dataset of LIB time-series and descriptive reports, consisting of 1,792 time-series samples, each comprising 259,778 minutes,  aligned with 72,520 reports, along with a benchmark suite covering three representative battery management tasks. All code and data are open-sourced.

\section{Results}

\subsection{Illustration of the TimeSeries2Report framework}
\begin{figure}[!t]
\centering
\includegraphics[width=\textwidth]{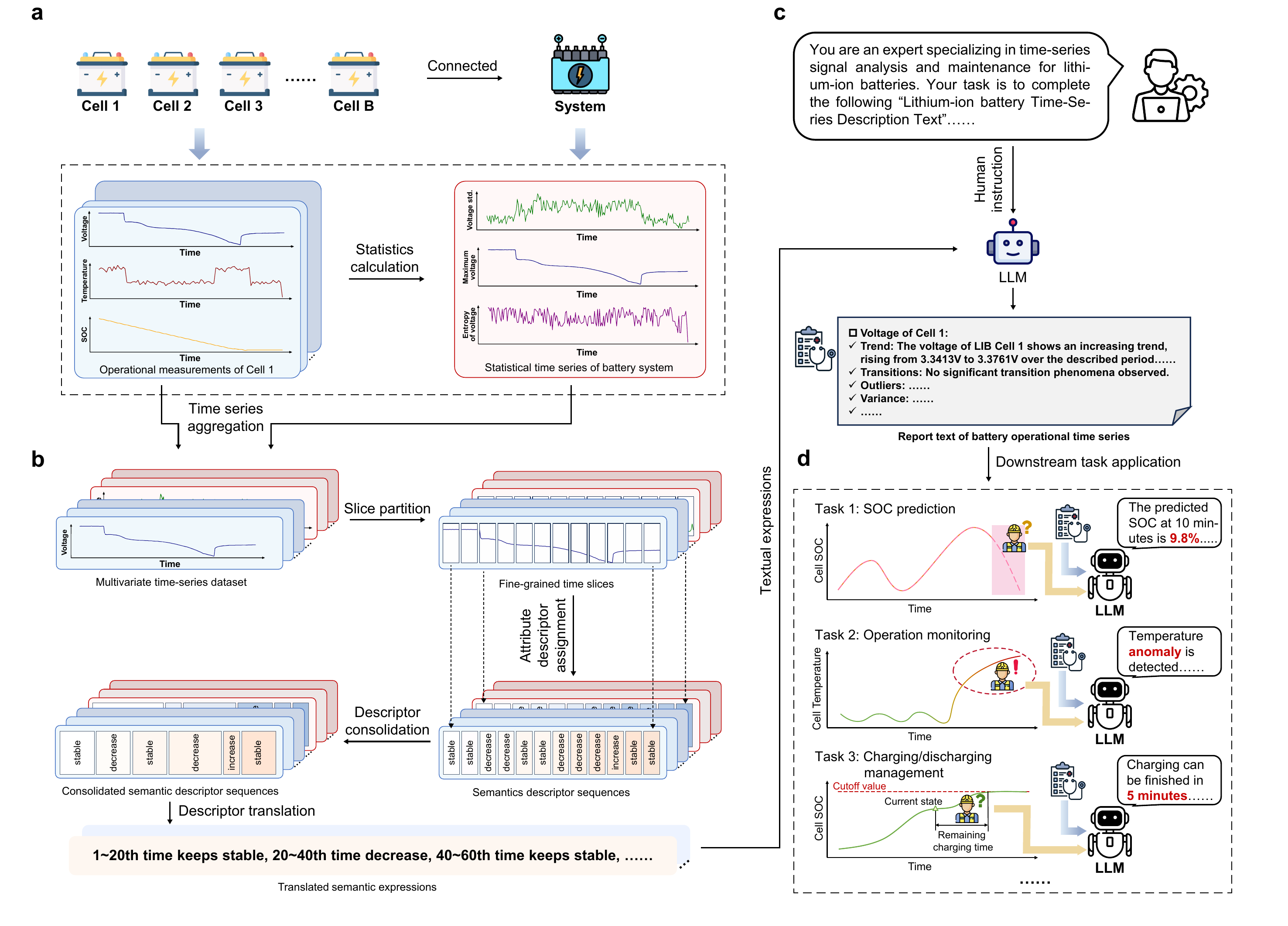}
\caption{Overview of the TimeSeries2Report framework. \textbf{a}, Data acquisition and system-level statistics. Operational time-series data are collected from all LIB cells arranged in series and/or parallel configurations. Cross-cell statistics, i.e., mean (avg.), standard deviation (std.), and entropy, are computed at each time stamp to capture system-level dynamics. \textbf{b}, Time-series to expression. Each time series is uniformly segmented into short temporal slices, and each slice is evaluated using multiple attributes (e.g., trend, fluctuation, transition) to generate local semantic descriptors such as increasing, decreasing, or stable. Adjacent slices with identical descriptors are merged to reduce redundancy, and the resulting consolidated descriptors are translated into standardized expressions. \textbf{c}, Expression to report. The LLM receives (\romannumeral1) the derived expressions, and (\romannumeral2) instructions specifying the report’s desired style and output structure. It integrates these inputs to synthesize a coherent, evidence-grounded report describing the operational behavior of the BESS. \textbf{d}, Downstream applications. The generated report serves as input for LLM-driven O$\&$M tasks, demonstrated here for three representative examples: SOC prediction, operational monitoring with anomaly detection, and charging/discharging management with interpretable decision support.}\label{fig_framework}
\end{figure}

As illustrated in the top panel of Fig. \ref{fig_framework}a, a typical BESS consists of numerous LIB cells connected in series and/or parallel configurations. The first stage of the TS2R framework involves data acquisition and pre-processing. For each cell, we record key operational measurements such as voltage, current, and temperature  and compute system-level statistics across all cells at each time. These aggregated metrics, including the mean, maximum, minimum, standard deviation, and entropy, characterize the aggregate behavior and heterogeneity of the BESS (bottom panel of Fig. \ref{fig_framework}a, Methods \ref{statistical-calculation}).

Subsequently, the time-series data from all individual cells and the corresponding system-level statistics are aggregated into a unified multivariate dataset. Each time series is then segmented into fine-grained, temporally ordered slices to capture short-term dynamical behavior. The segmentation window is chosen to be sufficiently short to preserve local trends while retaining meaningful temporal structure rather than noise \cite{kil1997optimum}. In this work, the temporal window width is fixed at 10 time stamps, determined through expert knowledge and representative observations (see Methods \ref{slice-partition}). To characterize these short-term dynamics, we design a set of attributes that comprehensively describe local temporal behavior (see Methods \ref{design-semantic-attribute} and Table \ref{tab1}). For each attribute of each slice, semantic descriptors are automatically assigned using a rule-based computational pipeline (see Methods \ref{assigning-rule} and Table \ref{tab-descriptor-assignment}). For example, the attribute “\textit{trend}’’ is mapped to semantic descriptors such as \textit{increasing}, \textit{stable}, or \textit{decreasing}. This procedure transforms raw numerical time series into sequences of semantic descriptors. Adjacent slices with identical semantic descriptors are then merged to reduce redundancy and enhance temporal coherence. This merging is performed across the full time span and all variables, yielding compact yet information-rich representations that retain the salient dynamical features of the system (see Methods \ref{consolidation-descriptor}). Finally, the consolidated descriptor sequences are translated into human-readable expressions using predefined phrasing rules (see Methods \ref{translating-descriptors}). At this stage (Fig. \ref{fig_framework}b), each time series is converted into a concise textual summary that integrates multiple dynamic attributes, thereby bridging quantitative time-series data with interpretable natural-language representations. Next, the framework synthesizes the generated expressions into a comprehensive report using an LLM (Fig. \ref{fig_framework}c). The LLM receives two inputs: (\romannumeral1) the textual expressions generated in the previous stage and (\romannumeral2) detailed instructions specifying the report’s desired descriptive style and output structure (see Methods \ref{convert-expression-into-report}). By integrating these inputs, the LLM produces a coherent, context-aware report that describes the original time-series data from multiple dynamic-attribute perspectives.

The generated report serves as readily usable inputs for downstream LLM-driven O$\&$M tasks. We demonstrate their applicability in three representative scenarios: SOC prediction, operational monitoring (abnormality detection), and charging/discharging management (Fig. \ref{fig_framework}d and Methods \ref{ref4-8}). Given these sentences and the task-specific user queries, the LLM generates interpretable responses supported by explicit, semantically grounded reasoning, thereby enhancing transparency and explainability throughout the battery O$\&$M workflow. See supplementary video \texttt{demo.mp4} for full demonstration.

\subsection{Benchmarking the performance of TimeSeries2Report on open-source datasets of standalone LIB cells}
To evaluate the performance of TS2R, we sourced two public datasets from Refs.~\cite{severson2019data} and \cite{zhu2022data}, and curated them into the MIT dataset and TJU dataset, respectively (see Supplementary Note 1, Table S1 and Fig. S1 for detailed curation steps). Both datasets consist of measurements from single, standalone LIB cells but differ in cell chemistry: the MIT dataset corresponds to lithium iron phosphate (LFP) cells, while the TJU dataset contains nickel cobalt aluminum oxide (NCA) cells. Since these datasets involve standalone cells rather than interconnected multi-cell systems, the computation of system-level statistics and downstream multi-cell operations is bypassed. Instead, we focus on evaluating the quality of reports generated for individual LIB cells across a range of operating conditions, including constant-current (CC) charging, CC-to-constant-voltage (CV) charging, idle-to-CC discharging, and CC discharging-to-idle transitions, as well as across chemistries (LFP and NCA) and LLM backbones (Qwen3-14B~\cite{yang2025qwen3}, Llama3.1-8B~\cite{dubey2024llama}, DeepSeek-v3.2~\cite{guo2025deepseek}, and ChatGLM-4.5~\cite{zeng2025glm}).

\begin{figure}[!t]
\centering
\includegraphics[width=\textwidth]{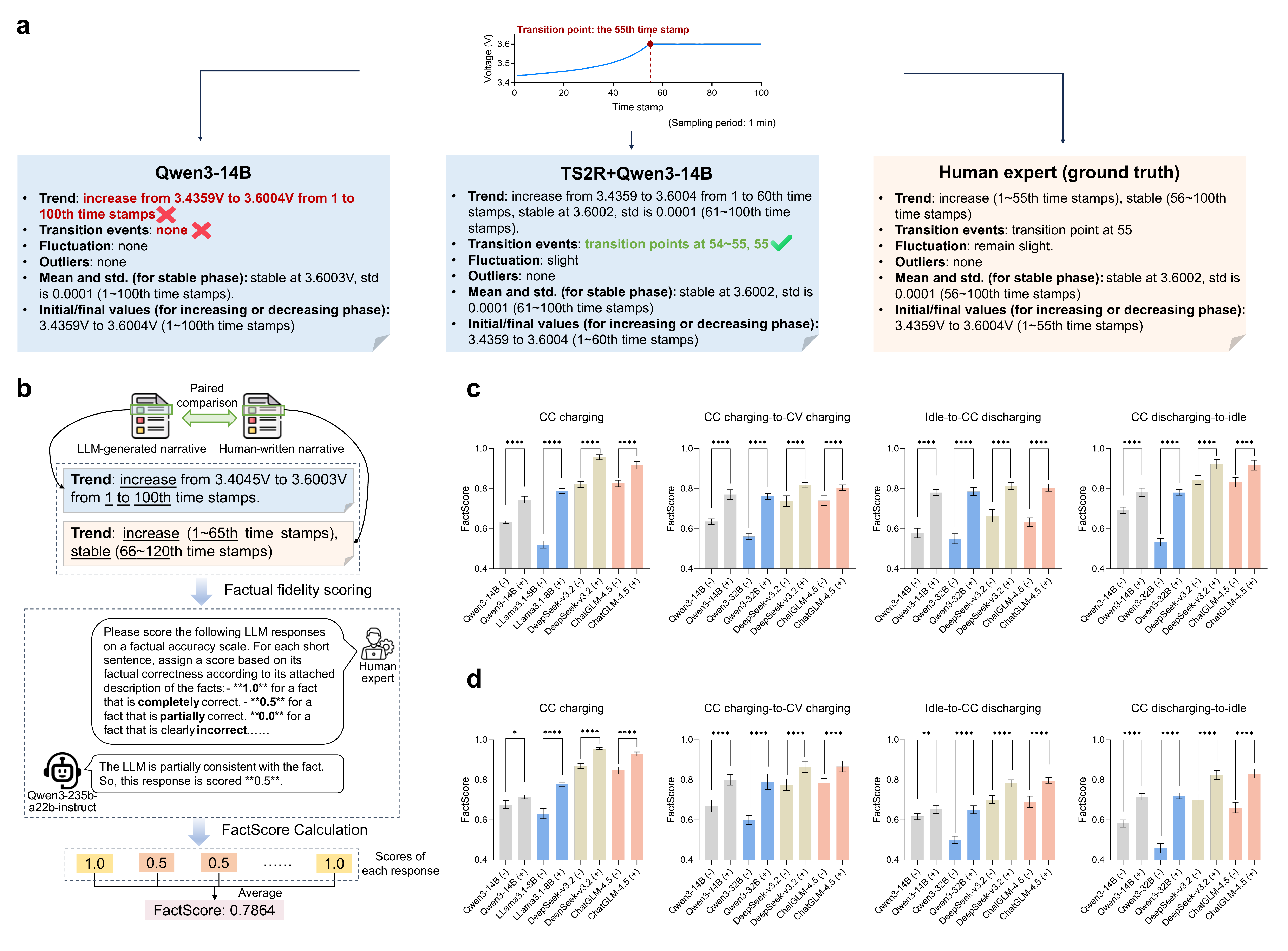}
\caption{\textbf{Quantitative evaluation of TS2R on public datasets of single, standalone LIB cells.} \textbf{a}, Top: Representative voltage time series from the MIT dataset showing a LIB cell undergoing CC-CV charging. Bottom: Comparison of reports generated by Qwen3‑14B prompted with (left) raw numerical time series, (middle) TS2R-parsed textual input, and (right) an expert-written reference. The TS2R-enhanced report correctly identifies the CC-CV phase transition at the $\sim$ 55$^\text{th}$ time stamp, whereas the baseline fails to do so. \textbf{b}, Schematic of the FactScore computation pipeline, which quantifies the factual consistency between generated reports and expert annotations. \textbf{c}, FactScore comparison for LLMs prompted with raw numerical time series (–) versus TS2R-parsed text (+), evaluated on the MIT dataset. \textbf{d}, Same as (\textbf{c}), evaluated on the TJU dataset. TS2R consistently improves performance across all models (Qwen3‑14B, Llama3.1‑8B, DeepSeek‑v3.2, ChatGLM-4.5) and conditions. DeepSeek‑v3.2 (+) and ChatGLM‑4.5 (+) achieve the highest overall performance. Bars represent mean values; error bars indicate 95$\%$ confidence intervals (CI); overlaid black dots show individual FactScore values. *: $p < 0.1$, **: $p < 0.01$, ****: $p < 0.0001$; $p$-values computed using a one-sided paired Wilcoxon test.}\label{figure-single-cell}
\end{figure}

In Fig. \ref{figure-single-cell}a, we present representative outputs based on the voltage time series of a single LIB cell, where the voltage increases continuously until the 55$^\text{th}$ time stamp and remains stable thereafter (top panel). The three columns compare: (\romannumeral1) the direct output of Qwen3‑14B prompted with raw numerical time-series data (left), (\romannumeral2) the output from TS2R-processed input using the same LLM (middle, TS2R+Qwen3‑14B), and (\romannumeral3) an expert-written reference report serving as ground truth (right). Unlike the raw-input baseline, TS2R+Qwen3‑14B accurately identifies both the global trend (initial increase followed by stability) and the transition point at the 55$^\text{th}$ time stamp. It also more consistently aligns with expert annotations across other descriptive attributes. In contrast, the baseline Qwen3-14B fails to capture these patterns reliably.

To quantify the generated report quality, we introduce FactScore \cite{min2023factscore}, a metric that assesses the alignment between generated reports and expert-written references (see Supplementary Note 2 for annotation protocol). As illustrated in Fig. \ref{figure-single-cell}b, each sentence generated for different signals (e.g., voltage, temperature, current) and attributes (e.g., trend, transition, outlier) is compared to its expert counterpart. An independent judging LLM (Qwen3-235B-A22B-Instruct-2507 \cite{yang2025qwen3}) evaluates each sentence pair and assigns a score based on consistency: 1 for high consistency, 0.5 for partial consistency, and 0 for low consistency. The final FactScore is computed as the average score across all attribute-level evaluations for a given time series. Full judging prompts are detailed in Supplementary Note 3.

We quantitatively assess the performance of TS2R using FactScore on both the MIT (Fig. \ref{figure-single-cell}c) and TJU (Fig. \ref{figure-single-cell}d) datasets, covering a range of operating conditions and LLM backbones, including Qwen3‑14B, Llama3.1‑8B, DeepSeek‑v3.2 and ChatGLM-4.5. To evaluate effectiveness, we compare two prompting strategies: (\romannumeral1) directly inputting raw numerical time-series data as prompts into the LLM, and (\romannumeral2) first converting the data into structured natural language using TS2R prior to prompting. Across all conditions, the TS2R-augmented approach consistently outperforms direct numerical prompting, yielding FactScore improvements of 6–57$\%$, most with statistical significance ($p < 0.0001$). These results highlight the inherent difficulty text-based LLMs face when interpreting raw time-series data, particularly in capturing complex temporal features such as trends, transitions, and outliers. By translating numerical signals into interpretable natural language, TS2R effectively unlocks the semantic reasoning capabilities of general-purpose LLMs.

The consistent gains observed across all four models underscore the model-agnostic nature of TS2R, demonstrating that its benefits generalize across architectures without requiring model-specific tuning. Importantly, TS2R achieves this performance without any retraining, offering a lightweight, scalable, and adaptable solution for generating descriptive reports from single-cell time-series data. These findings position TS2R as a practical and interpretable framework for real-world battery management applications.

\subsection{Evaluating TS2R in real-world, system-level settings}
\begin{figure}[!t]
\centering
\includegraphics[width=\textwidth]{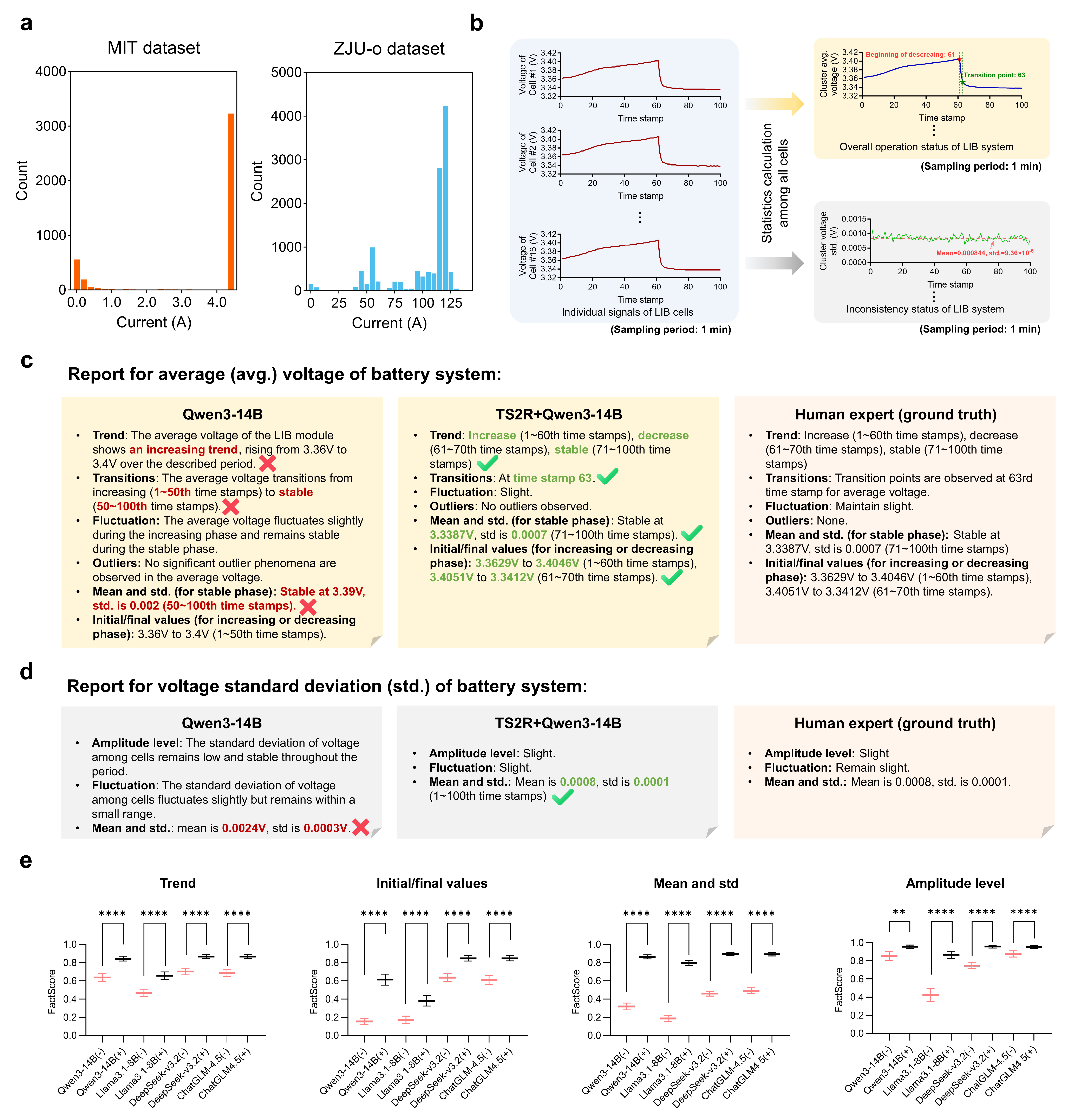}
\caption{\textbf{Performance evaluation of TS2R on a real-world BESS dataset.} \textbf{a}, Distribution of discharging currents from the MIT dataset (lab-scale) and the ZJU-o dataset (real-world). The ZJU-o distribution exhibits a markedly broader range, highlighting the greater variability and complexity of LIB operating conditions in field deployments. \textbf{b}, Schematic illustration of the computation of system-level statistical time series, including the average and standard deviation of voltage across 16 cells per module. \textbf{c}–\textbf{d}, Example system-level reports generated by Qwen3-14B with raw numerical input (left), TS2R-processed semantic input (middle), and expert-written references (right), based on the voltage mean (\textbf{c}) and standard deviation (\textbf{d}) time series shown in (\textbf{b}). With raw numerical input, Qwen3-14B misinterprets trends and fails to detect key transitions, whereas TS2R enables accurate identification of a sharp voltage drop and the corresponding transition point. \textbf{e}, FactScore evaluation on the ZJU-t dataset assessing system-level report quality across four attributes (trend, initial/final values, temporal mean and standard deviation and amplitude level) and four LLM backbones (Qwen3-14B, Llama3.1-8B, DeepSeek-v3.2, ChatGLM-4.5). Across all settings, LLMs prompted with TS2R-derived semantic input substantially outperform those given raw numerical data. Horizontal lines indicate mean FactScore; error bars represent 95$\%$ CI. **: $p < 0.01$; ****: $p < 0.0001$ (one-sided paired Wilcoxon test).}\label{figure-system-level}
\end{figure}

While the MIT and TJU datasets were collected under well-controlled laboratory conditions and focus exclusively on standalone LIB cells, evaluating TS2R in a realistic, system-level context requires more representative field data. To this end, we collected operational data from a deployed BESS consisting of 28 modules, each containing 16 LFP cells connected in series. Over a six-month period (December 1, 2023 to May 30, 2024), we recorded minute-level time-series measurements of voltage, temperature, and SOC from each individual cell, as well as module-level current. Since all cells within a module are connected in series, the module current is equivalent to the current flowing through each cell. In total, this yielded 28 modules × 16 cells × 4 variables $=$ 1,792 time series, each comprising 259,778 minutes. The dataset includes over 400 operational charge-discharge cycles, executed under constant-power (CP) protocols in response to dynamic grid demands. Each sample in this dataset, referred to as the ZJU‑o dataset (with “o” denoting omni), consists of synchronized time-series data for voltage, temperature, SOC, and current, capturing real-world, module-level battery behavior under diverse operating conditions (see Supplementary Note 4, Table S2 and Fig. S2 for data acquisition details). The difference in operational complexity between lab-scale and real-world deployments is illustrated in Fig. \ref{figure-system-level}a and Fig. S3, which compare the distributions of charging and discharging currents in the MIT and ZJU‑o datasets. The ZJU‑o dataset exhibits substantially broader current variability, reflecting increased operational diversity and environmental uncertainty – factors that pose a more rigorous test for evaluating the robustness of TS2R-generated reports. Due to the sheer scale of the ZJU‑o dataset, manual annotation by human experts is impractical. To enable benchmark evaluation, we curated a smaller test set referred to as the ZJU‑t dataset (with “t” for test). This was constructed by: (\romannumeral1) randomly sampling a module index $i$, (\romannumeral2) selecting a random initial time $t$ within the operational timeline, and (\romannumeral3) recording all four variables (voltage, temperature, SOC, current) for all 16 cells in module $i$ over the window $\left[t,t+100\right]$ minutes. This process was repeated 100 times, resulting in 100 time-series samples with expert-written annotations generated using the same protocol described in Supplementary Note 2. The ZJU‑t dataset captures module-level behavior by including multiple cells per module, allowing us to compute system-level statistics (note: while technically “module-level,” we use “system-level” hereafter without ambiguity). As shown in Fig. \ref{figure-system-level}b, we aggregate per-cell measurements at each time stamp to compute summary statistics (e.g., mean and standard deviation of voltage), yielding system-level features that reflect both aggregate trends and inter-cell variability. This enables TS2R to operate hierarchically, interpreting battery behavior at both the individual cell and system level, thereby supporting hierarchical interpretation of battery system behavior.

In Fig. \ref{figure-system-level}c, we present a representative report generated from the system-level voltage statistics shown in Fig. \ref{figure-system-level}b. In this example, the average voltage exhibits a transition from a stable increase to a sharp decrease around the 61$^\text{st}$$\sim$63$^\text{rd}$ time stamp. When prompted with raw numerical data, the LLM (Qwen3‑14B) fails to capture this transition and incorrectly reports a monotonic increase over the entire period. In contrast, the same LLM, when provided with TS2R-parsed input (TS2R+Qwen3‑14B), correctly identifies the increasing phase, the sharp decrease, and the subsequent stabilization. The resulting report closely matches the expert-written reference in both qualitative trend description and quantitative accuracy. A similar effect is observed in Fig. \ref{figure-system-level}d, where the LLM is prompted with the standard deviation of voltage: TS2R again enables the model to detect and articulate key statistical transitions that are overlooked when using raw numerical input alone.

To systematically assess performance, we applied the same evaluation protocol used in Fig. \ref{figure-single-cell}b to the ZJU‑t dataset, measuring report quality at both the single-cell and module (system) levels. Fig. \ref{figure-system-level}e and Fig. S4a summarize the FactScore results for module-level reports, evaluated across nine descriptive attributes shown in Table \ref{tab-attribute}, three operational states (charging, discharging, and idle), and four LLM backbones (Qwen3‑14B, Llama3.1‑8B, DeepSeek‑v3.2 and ChatGLM-4.5). Corresponding FactScore results for individual cells are provided in Supplementary Fig. S4b. Across all conditions, LLMs augmented with TS2R consistently outperform those using raw numerical input, except in cases where the baseline performance is already near saturation (for example, the fluctuation attribute in Fig. S4a), demonstrating significantly improved accuracy in capturing module-level dynamics. Among the tested models, both TS2R+DeepSeek‑v3.2 and TS2R+ChatGLM-4.5 achieve the remarkably promising performance, with FactScore metrics approaching 0.8$\sim$1.0 in most cases. These results confirm that TS2R substantially enhances LLMs’ ability to interpret complex, real-world time-series data and generate reports that align closely with expert-written references, both at the single-cell and system levels. This underscores the potential of TS2R to advance downstream battery O$\&$M tasks through accurate, interpretable, and scalable report generation.

\subsection{TS2R enhances the accuracy of SOC prediction}
We next demonstrate how TS2R can be integrated into and improve the performance of a critical downstream task in LIB management: SOC prediction, which is essential for understanding the charging and discharging status.

\begin{figure}[!t]
\centering
\includegraphics[width=\textwidth]{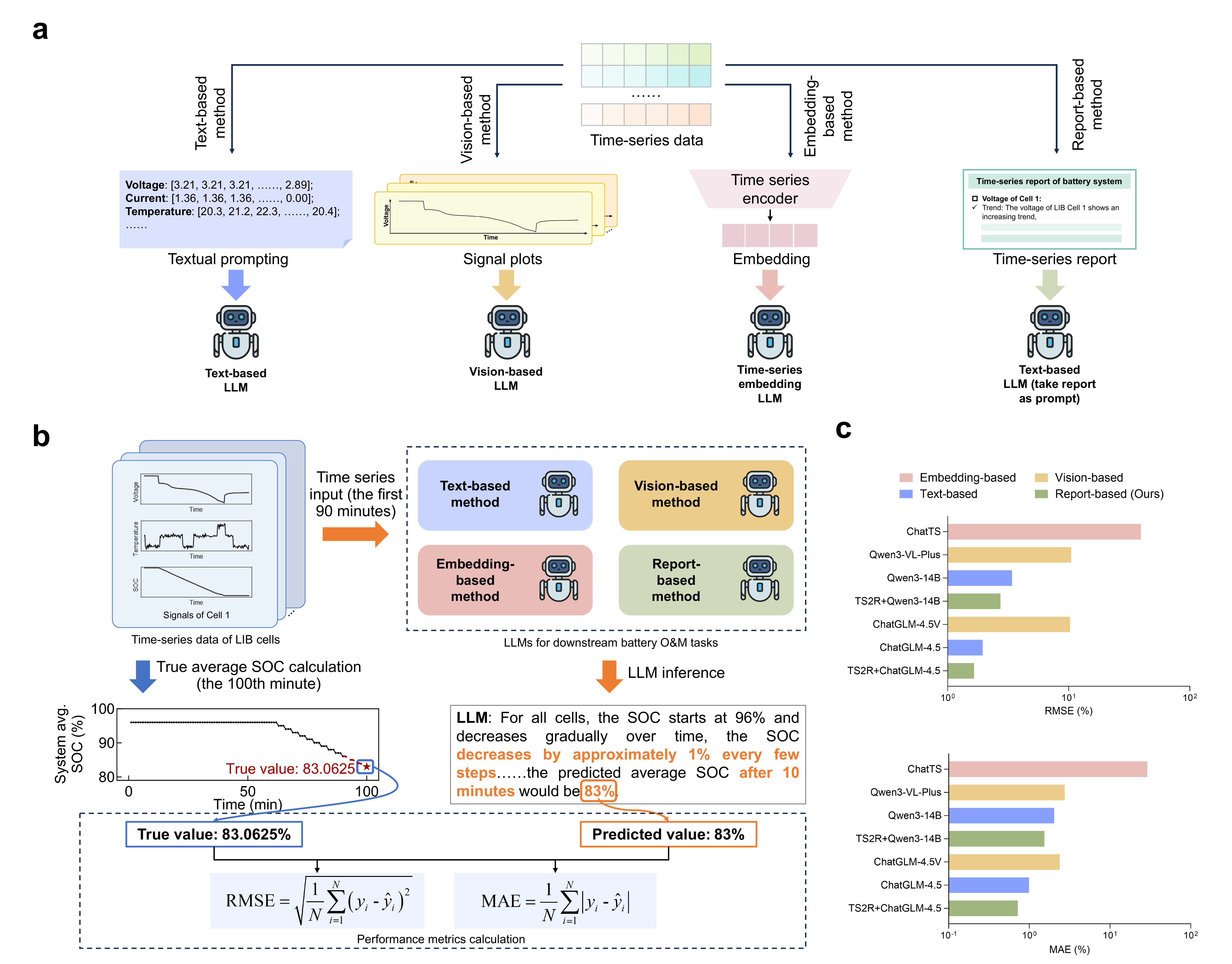}
\caption{\textbf{TS2R improves the accuracy of SOC prediction.} \textbf{a}, Illustration of four categories of LLM-based methods: (\romannumeral1) Text-based LLMs, which take raw numerical time-series data as text input; (\romannumeral2) Vision-based LLMs, which process time-series plots as images; (\romannumeral3) Time-series embedding models, which encode time-series data into deep vector embeddings aligned with language tokens; and (\romannumeral4) Report-based LLMs, where TS2R converts time-series into natural language reports used as prompts. \textbf{b}, Evaluation pipeline for SOC prediction. Using the ZJU-t dataset, each method is prompted to predict the system-level average SOC at minute 100, based on time-series input up to minute 90. Predictions are compared against ground truth, and performance is quantified RMSE and MAE. \textbf{c}, SOC prediction accuracy of seven LLM-based methods, shown as bar plots for RMSE and MAE. Report-based LLMs (TS2R+Qwen3‑14B, TS2R+ChatGLM‑4.5) consistently achieve the lowest prediction errors, demonstrating the effectiveness of TS2R in enhancing temporal reasoning.}\label{figure-soc-prediction}
\end{figure}

To ensure a fair comparison, we focus exclusively on LLM-based approaches, as outlined in Fig. \ref{figure-soc-prediction}a. The methods are grouped into four categories: (\romannumeral1) Text-based LLMs (Qwen3‑14B, ChatGLM‑4.5), which take raw numerical time-series values as direct textual input. (\romannumeral2) Vision-based LLMs (Qwen3‑VL‑Plus~\cite{yang2025qwen3}, ChatGLM‑4.5V~\cite{hong2025glm}), which interpret time-series data rendered as graphical plots. (\romannumeral3) Time-series embedding-based models (ChatTS \cite{xie2024chatts}), which encode numerical time series into dense vector representations aligned with language tokens. (\romannumeral4) Report-based LLMs (TS2R+Qwen3‑14B, TS2R+ChatGLM‑4.5), which leverage TS2R to convert raw time-series data into structured natural-language reports. DeepSeek-v3.2 and Llama3.1-8B are excluded from the following analyses, as they do not have vision-based counterparts.

To evaluate the performance of the seven models across the four categories, we followed the assessment pipeline shown in Fig. \ref{figure-soc-prediction}b. Using the ZJU-t dataset (see Fig. \ref{figure-system-level}e), we assessed prediction accuracy using two standard metrics: root mean squared error (RMSE) and mean absolute error (MAE). For each test sample, the first 90 minutes of the time series were provided as input, and each method was tasked with predicting the system-level average SOC at minute 100. RMSE and MAE were then calculated by comparing the predicted SOC with the ground truth at that time point. Prompting configurations for each method are detailed in Supplementary Note 5.

Example outputs from the seven SOC prediction methods are shown in Fig. S5. In Fig. \ref{figure-soc-prediction}c, we compare their prediction accuracy across the four LLM-based categories. Among these, the time-series embedding model (ChatTS) exhibits the lowest accuracy. This is likely due to its training on a highly curated dataset composed primarily of synthetically generated sequences, which limits its generalization to real-world data such as the ZJU-t dataset that features complex and diverse latent temporal patterns. Although retraining ChatTS on ZJU-t could improve its performance, doing so would require substantial computational resources and time. This limitation highlights a key advantage of TS2R: it enhances LLM performance without the need for retraining. Vision-based LLMs (Qwen3‑VL‑Plus and ChatGLM‑4.5V) rank second lowest in performance. While transforming numerical time series into visual plots helps preserve coarse trends, the resulting pixel-based representations introduce quantization artifacts that constrain the model’s ability to perform fine-grained reasoning, thereby reducing prediction precision. Text-based LLMs (Qwen3‑14B and ChatGLM‑4.5) perform moderately well, producing acceptable SOC predictions when directly prompted with raw numerical input. In contrast, report-based LLMs, which utilize TS2R-generated natural language descriptions, consistently achieve the highest prediction accuracy. Notably, the TS2R+Qwen3‑14B model reduces RMSE by 20$\%$ and 74$\%$ compared to its corresponding text-based and vision-based variants (Qwen3‑14B and Qwen3‑VL‑Plus, respectively). Similar improvements are observed with the ChatGLM-based models. These results underscore the effectiveness of TS2R in enabling LLMs to better interpret complex time-series data and perform accurate forecasting without the need for architecture modification or retraining.

\subsection{TS2R enables accurate detection of abnormalities in LIB operation monitoring}
\begin{figure}[!t]
\centering
\includegraphics[width=\textwidth]{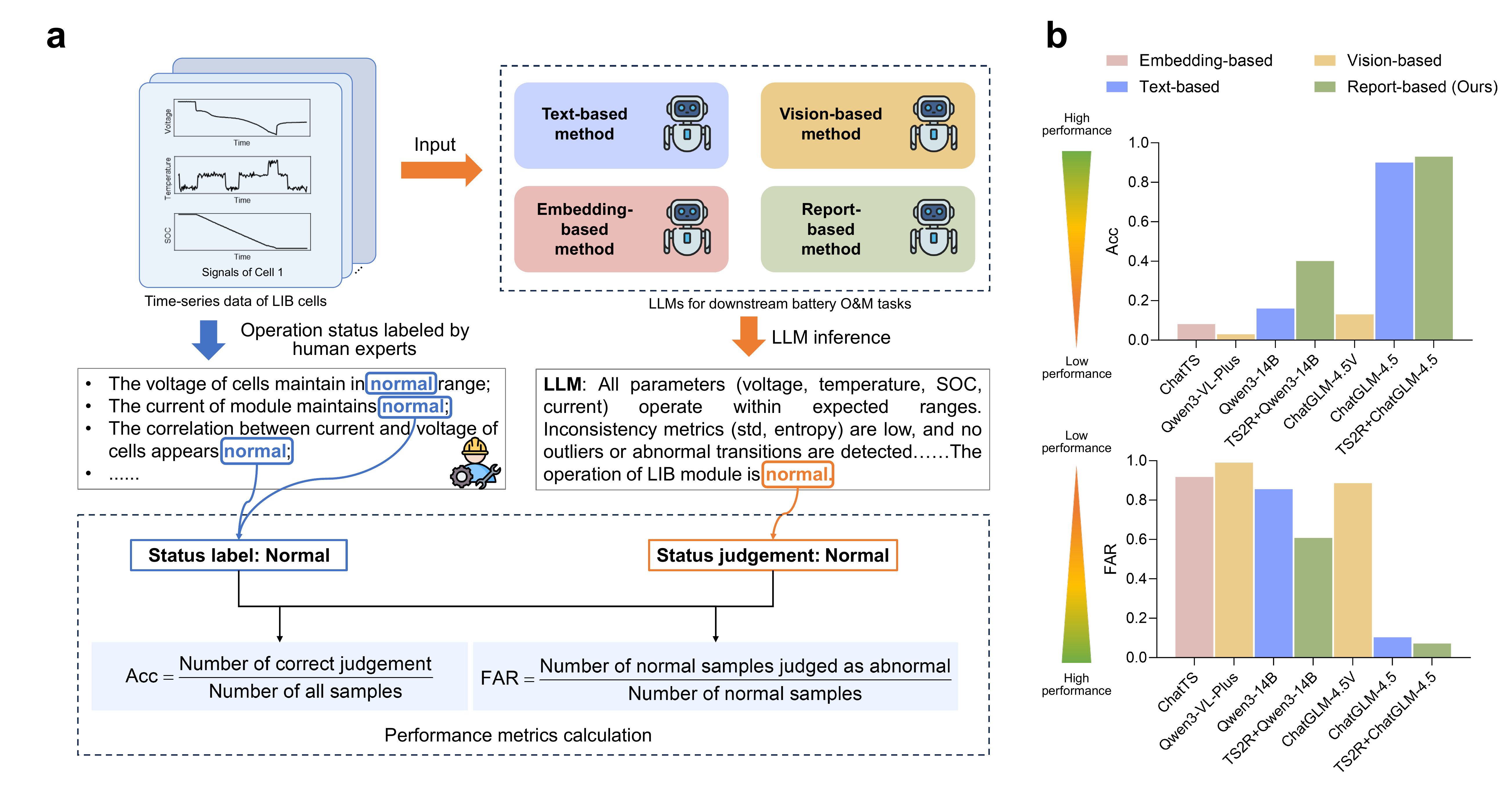}
\caption{\textbf{TS2R enables accurate abnormality detection in LIB operation monitoring.} \textbf{a}, Quantitative evaluation pipeline for anomaly detection using the seven LLM-based methods across four categories described in Fig. 4. Each method is prompted to classify whether a time-series sample is normal or abnormal. If labeled as abnormal, the model is further queried to identify the type of anomaly (e.g., voltage or temperature-related). Performance is assessed using two metrics: Acc and FAR. \textbf{b}, Comparison of the seven methods on the ZJU-ta dataset, which includes three curated abnormal scenarios and reflects the real-world rarity of anomalies. Report-based methods (TS2R+ChatGLM-4.5 and TS2R+Qwen3‑14B) consistently outperform other approaches, achieving the highest accuracy and lowest FAR within their respective LLM families. TS2R+ChatGLM-4.5 achieves the best overall performance, demonstrating its suitability for real-time LIB operation monitoring.}\label{figure-anomaly-detection}
\end{figure}

A fundamental task in LIB operation monitoring is the real-time detection of abnormalities, enabling timely interventions to prevent safety hazards and performance degradation. The goal of anomaly detection is twofold: (\romannumeral1) to accurately identify true abnormal events, and (\romannumeral2) to minimize false positives, i.e., to avoid misclassifying normal behavior as abnormal. We evaluated this task using the same seven LLM-based methods described earlier. Each method was first prompted to classify a given time series as either normal or abnormal. If an abnormality was detected, the model was further queried to specify the anomaly type (e.g., voltage-related, temperature-related). Prompting details are provided in Supplementary Note 5. Since the original ZJU-t dataset contains only normal operating conditions, it is insufficient for this evaluation. To address this, we constructed the ZJU-ta dataset (“a” indicating the inclusion of abnormal cases) by replacing the final three normal-operation samples in ZJU-t with three curated abnormal scenarios: (1) temperature abnormality in a single cell, (2) voltage abnormality in a single cell, and (3) voltage abnormality at the system level. Details of these abnormal cases are provided in Table S4. Importantly, the ratio of abnormal to normal samples reflects the low incidence of anomalies observed in real-world deployments. The evaluation pipeline is illustrated in Fig. \ref{figure-anomaly-detection}a. For each sample, all methods were prompted to perform anomaly detection. Predictions were compared against expert annotations (ground truth), and two metrics were computed over the ZJU-ta dataset: Accuracy (Acc), measuring the ability to correctly identify true anomalies (objective \romannumeral1), and False Alarm Rate (FAR), quantifying the frequency of incorrectly labeling normal behavior as abnormal (objective \romannumeral2).

Example outputs for both abnormal and normal samples across the seven LLM-based methods are presented in Figs. S6 and S7. Fig. \ref{figure-anomaly-detection}b summarizes the quantitative comparison using two evaluation metrics Acc and FAR. Among all methods, the time-series embedding-based model (ChatTS) achieves an accuracy below 0.1 and FAR around 0.9, clearly inadequate for practical LIB operation monitoring. Interestingly, we observe significant performance differences even between models with the same LLM backbone (e.g., Qwen3-series vs. ChatGLM-series), likely due to architectural differences inherent to the base models. However, since such model-specific factors are not the focus of this study, we compare methods within each LLM family. For the Qwen3-series, the vision-based method (Qwen3-VL-Plus) achieves the lowest accuracy ($\sim$0.03), while the report-based method (TS2R+Qwen3-14B) improves accuracy by more than a factor of two over the text-based variant (Qwen3-14B), and reduces FAR by 28.92$\%$. Similar trends are observed within the ChatGLM series. Notably, TS2R+ChatGLM-4.5 achieves the best overall performance, with an accuracy of approximately 0.9 and FAR below 0.1, making it the most suitable choice for real-world LIB operation monitoring. These findings are further supported by the qualitative comparison of model outputs for the three abnormal samples, detailed in Table S5.

\subsection{TS2R supports more precise predictive decision-making for LIB charging and discharging management}
\begin{figure}[!t]
\centering
\includegraphics[width=\textwidth]{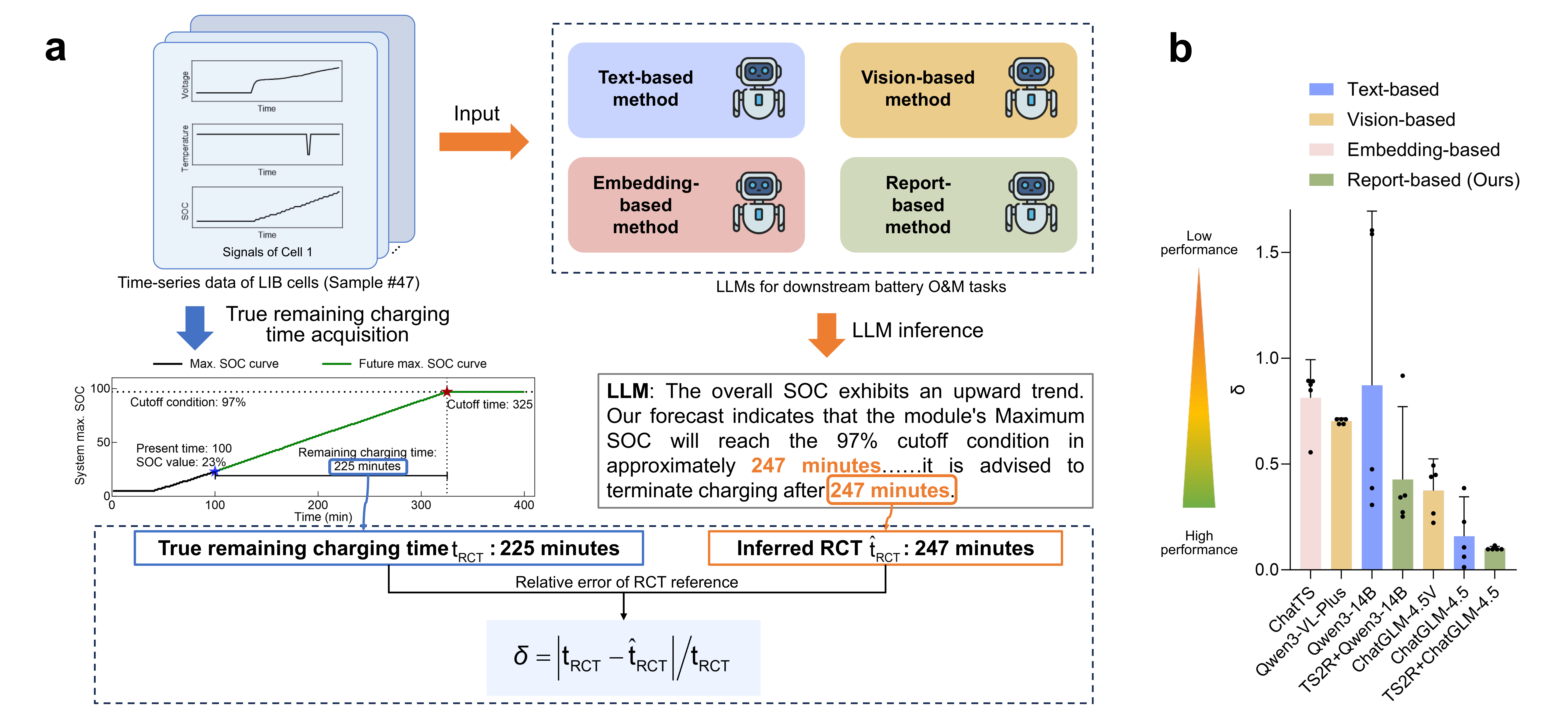}
\caption{\textbf{TS2R supports predictive decision-making for charging management in LIB systems.} \textbf{a}, Evaluation protocol for estimating the RCT. Each method is prompted to predict how long it will take for the system-level maximum SOC to reach the manufacturer-defined cutoff (97$\%$), given only time-series data up to minute 100. Ground-truth RCT is obtained from the ZJU-o dataset, and prediction accuracy is measured using the relative error $\delta={\left|t_\text{RCT}-\hat{t}_\text{RCT} \right|}/{t_\text{RCT}}$. \textbf{b}, Comparison of relative errors $\delta$ across seven LLM-based methods. To account for output stochasticity, predictions are repeated five times per method. Report-based methods (TS2R+Qwen3‑14B, TS2R+ChatGLM‑4.5) consistently achieve the lowest relative errors and tightest confidence intervals. TS2R+ChatGLM‑4.5 exhibits the highest accuracy and consistency, underscoring its potential for deployment in real-world BESS management.}\label{figure-management}
\end{figure}

Since TS2R has demonstrated strong predictive capabilities, we next investigate whether these benefits can be translated into improved decision-making. A key function of BESS management is to make timely decisions to terminate charging or discharging in order to prevent over-stressing or under-stressing individual cells – actions critical for both safe operation and long-term longevity of the system. Typically, BESS managing system uses the system-level maximum or minimum SOC, i.e., the highest or lowest SOC across all LIB cells, as the criterion to determine safe charging/discharging cutoffs. To illustrate how TS2R supports predictive decision-making for charging and discharging management, we apply two filtering conditions to the ZJU-t dataset: (\romannumeral1) the system must be mid-way through a charging cycle at minute 100, and (\romannumeral2) the corresponding ZJU-o dataset must show that the system-level maximum SOC continues to rise after minute 100, eventually reaching 97$\%$, a cutoff defined by the BESS manufacturer. This second condition ensures that discharging events (e.g., grid demand response) do not interfere, allowing the true remaining charging time (RCT) to be recorded as ground truth for evaluating predictive accuracy. Using these constraints, we identify a representative sample (Sample $\#$47) from the ZJU-t dataset, shown in Fig. \ref{figure-management}a. In this sample, the system-level maximum SOC at minute 100 is 23$\%$ (marked by a blue star in Fig. \ref{figure-management}a). The task is to predict the remaining charging time $\hat{t}_\text{RCT}$: how long it will take for the maximum SOC to reach the cutoff value of 97$\%$, given only time-series data up to minute 100. By referencing the corresponding ZJU-o record, the ground truth RCT is $t_\text{RCT}=225$ minutes (marked by a red star in Fig. \ref{figure-management}a). Prediction performance is quantified by the relative error $(\delta={\left|t_\text{RCT}-\hat{t}_\text{RCT} \right|}/{t_\text{RCT}}) $. A smaller $\delta$ indicates a more accurate predictive decision by the LLM-based method. Details of the prompting setup are provided in Supplementary Note 5. Fig. S8 presents sample predictions of RCT $\hat{t}_\text{RCT}$ from the seven LLM-based methods.

Because LLMs are inherently generative and exhibit stochastic outputs, we repeated the relative error computation five times for each LLM-based method, following the evaluation protocol outlined in Fig. \ref{figure-management}a. In Fig. \ref{figure-management}b, we compare the resulting relative errors $\delta$ for RCT predictions across the seven methods. The embedding-based model ChatTS yields substantially higher $\delta$ values than either report-based method (TS2R+Qwen3‑14B and TS2R+ChatGLM‑4.5). Within each LLM family, the report-based variant consistently achieves the lowest relative error and tighter confidence intervals, indicating superior prediction accuracy and consistency. This demonstrates TS2R’s ability to enhance both the precision and reliability of LLM-based predictive decision-making. Notably, the combination of TS2R and ChatGLM‑4.5 delivers the most accurate and consistent predictions overall, with a remarkably low $\delta$ and narrow confidence band. These results underscore its strong potential for real-world deployment in BESS management systems.

\section{Discussion}\label{discussion}
In summary, this work presents TimeSeries2Report, a prompting framework that enables LLMs to effectively interpret and manage BESS by transforming raw multivariate time-series data into structured, semantically enriched natural language reports. By bridging low-level operational measurements with high-level contextual understanding, TS2R enhances the interpretability and decision-making capability of general-purpose LLMs. Through extensive benchmarking on both lab-scale (MIT, TJU) and real-world datasets (ZJU-t), we demonstrate that TS2R substantially improves the factual consistency of generated reports across diverse BESS operating conditions (CC charging, CC charging-to-CV charging, idle-to-CC discharging, and CC discharging-to-idle) and across multiple LLM backbones (Qwen3-14B, Llama3.1-8B, ChatGLM-4.5, DeepSeek-v3.2), outperforming baselines that rely on raw numerical input.

Among all tested combinations, TS2R+ChatGLM-4.5 and TS2R+DeepSeek-v3.2 achieve the highest factual accuracy and consistency. Notably, given the overwhelming volume of BESS operational data, expert annotation at scale is prohibitively time-consuming, creating a major bottleneck in developing domain-specific machine learning models. The high-quality outputs generated by TS2R+ChatGLM-4.5 and TS2R+DeepSeek-v3.2 enable automatic annotation of large-scale BESS data, providing a viable path toward scalable dataset construction. Leveraging this capability and accounting for API-call costs, we use TS2R+DeepSeek-v3.2 to annotate the entire ZJU-o dataset, making it, to the best of our knowledge, the first large-scale, paired BESS operational dataset with natural language annotations. We release this resource to support future research in LLM-driven BESS monitoring, prediction, and autonomous decision-making.

We further demonstrate that TS2R’s ability to generate reports with high factual accuracy and consistency directly supports improved performance across a wide range of downstream O$\&$M tasks, including SOC prediction, anomaly detection, and charging/discharging management. In all these tasks, TS2R outperforms competing approaches that represent time-series data as plots, embeddings, or plain numerical text. These findings reinforce TS2R’s core motivation: enhancing LLM comprehension of complex data by translating it into natural language. Among all evaluated combinations, TS2R+ChatGLM-4.5 consistently achieves the best trade-off between predictive accuracy, interpretability, and robustness across tasks. This highlights its strong potential for real-world deployment in intelligent, adaptive, and trustworthy BESS management systems.

One of the most notable advantages of TS2R is its ability to enhance LLMs’ capacity to interpret complex time-series data without requiring retraining, fine-tuning, or architectural modifications – processes that are often computationally expensive and time-consuming. Because TS2R operates purely as a prompting framework, it can leverage off-the-shelf LLMs via standard API calls, offering a lightweight, scalable, and interpretable solution for battery O$\&$M that is readily deployable across diverse systems. While TS2R already achieves strong performance in zero-shot settings, pairing it with fine-tuned LLMs could further improve results by internalizing domain-specific knowledge for battery time-series reporting. In such a supervised fine-tuning setting, the LLM is updated using available time-series data and corresponding offline reports. With the time-series input, human instruction and semantic descriptors serving as prompts, the model is trained to align its output with the reference report. The fine-tuning procedure is described as follows. Let $\mathbf{X}\in\mathbb{R}^{L\times V}$ denote the input multivariate battery time-series data, where $L$ represents the number of time stamps, and $V$ is the number of variables at each time stamp. First, TS2R pipeline converts time-series data into semantic descriptors:
\begin{equation}
    \mathcal{T}_\text{des}=f_\text{des}\left(\mathbf{X}\right)
\end{equation}
where $\mathcal{T}_\text{des}$ is the generated descriptors, $f_\text{des}$ denotes the conversion pipeline. Then TS2R takes the human instruction (denoted as $\mathcal{T}_\text{human}$) and the semantic descriptors as the input of LLM, and generates the corresponding time-series report $\mathcal{T}_\text{report}$:
\begin{equation}
    \mathcal{T}_\text{report}=f_\text{LLM}\left(\mathcal{T}_\text{des},\mathcal{T}_\text{human},\theta_\text{LLM}\right)
\end{equation}
where $f_\text{LLM}$ represents the function of LLM, and $\theta_\text{LLM}$ denotes its network parameters as the target of fine-tuning. The training objective follows a standard cross-entropy loss at the token level:
\begin{equation}\label{fine-tuning}
    \mathcal{L}=-\sum_{t=1}^T \sum_{k=1}^K y_{t,k}\text{log}\hat{y}_{t,k}
\end{equation}
where $T$ is the length of the LLM's generated report in tokens, $K$ is the vocabulary size, $y_{t,k}$ represents the one-hot encoded target probability for the $k^\text{th}$ token at the $t^\text{th}$ position in the report, $\hat{y}_{t,k}$ denotes the predicted probability of the $k^\text{th}$ token at the $t^\text{th}$ position, as produced by the LLM.

For more advanced applications, TS2R can be integrated into a retrieval-augmented generation (RAG) framework to support long-context and domain-enriched decision-making. In this setup, generated reports are segmented (e.g., system-level vs. cell-level content), indexed, and stored in a vector database. When a user issues a query, the most relevant report segments are retrieved and passed to the LLM as external knowledge. This enables the LLM to formulate accurate, context-aware responses, especially valuable in large-scale LIB systems where a single report may be too long for one-shot processing. Additionally, domain knowledge (e.g., safety thresholds, manufacturer specs) can be seamlessly integrated into the RAG database to further support professional-grade O$\&$M tasks.

\section{Methods}\label{method_section}
\subsection{Derivation of system-level statistical time series}\label{statistical-calculation}
The first step of the TS2R framework is to compute system-level statistical time series from individual LIB cell measurements. At each time point, we calculate a set of statistics, including mean, maximum, minimum, standard deviation, and Shannon entropy, for every measured variable across all cells in the system. In the MIT dataset, these variables include voltage, current, temperature, and charge/discharge capacities; in the TJU dataset, they include voltage, current, charge capacity, and discharge capacity; and in the ZJU-series dataset, they include voltage, temperature, SOC, and current. These statistics characterize both the aggregate behavior and the heterogeneity of the battery system, providing a compact representation of its spatiotemporal dynamics. The resulting system-level features, together with the raw time-series data from individual cells, constitute the preprocessed inputs for the subsequent stages of the TS2R pipeline.

\subsection{Partitioning time series into slices}\label{slice-partition}
Before semantic descriptors can be assigned, the raw time series must be decomposed into units that capture local, phase-specific behavior. Assigning a single descriptor to an entire sequence is typically inadequate, as LIB signals often exhibit complex, multi-stage dynamics. For example, a single time series may include multiple charging, discharging, and idle phases, making global labels such as \textit{increasing}, \textit{decreasing}, or \textit{stable} insufficient to characterize the full trajectory. To address this, we partition each time series into multiple equal-length time slices. The slice length is chosen to be sufficiently short to preserve local dynamical patterns, ensuring that each segment reflects a relatively coherent behavior that can be uniquely and accurately characterized by the semantic attributes introduced in the next step.

\subsection{Design of semantic attributes}\label{design-semantic-attribute}
To translate time-series data into natural language, we define a set of semantic attributes that specify the key informational elements to be captured in the generated report. These attributes determine which aspects of the time series should be described, such as trends, transitions, variability, and extremal values, and serve as the foundational units for producing interpretable and consistent expressions. The same semantic attribute framework is applied uniformly to both single-cell time series and system-level statistical time series, ensuring consistency across hierarchical levels of LIB monitoring.

To characterize both the behavior of individual cells (e.g., voltage, temperature) and the aggregate behavior of the entire system (mean, maximum, minimum for each variable across cells), we organize the semantic attributes into two categories: shape attributes and statistical attributes. Shape attributes capture temporal dynamics, including \textit{trend}, \textit{transition}, \textit{fluctuation}, and \textit{outliers}. Statistical attributes provide global summaries of each time series through metrics such as \textit{temporal mean}, \textit{standard deviation}, and \textit{initial and final values}. For time series reflecting inter-cell heterogeneity, such as system-level standard deviation and Shannon entropy, we apply the same two attribute categories but tailor their specific metrics. In this context, shape attributes include \textit{amplitude level} and \textit{temporal fluctuations}, while statistical attributes again consist of \textit{temporal mean} and \textit{standard deviation}. A complete description of the designed attributes is provided in Table \ref{tab1}. Notably, the framework is flexible and extensible: additional attributes can be seamlessly incorporated to accommodate specific configurations or operational requirements of different LIB systems.

\begin{table}[h]
\centering
\caption{Designed semantic attributes for single-cell and system-level statistical time series.}\label{tab1}\vspace*{1em}

\setlength{\tabcolsep}{6pt}
\renewcommand{\arraystretch}{1.25}

\begin{adjustbox}{width=\textwidth, center, keepaspectratio}
\fontsize{14pt}{20pt}\selectfont
\begin{tabular}{c | c | c | *{5}{>{\centering\arraybackslash}p{2.8cm}} | *{4}{>{\centering\arraybackslash}p{2.4cm}}}
\hline
\multicolumn{3}{c|}{\multirow{2}{*}[-18pt]{\makecell{Time Series}}}  & \multicolumn{5}{c|}{\makecell{Shape}} & \multicolumn{4}{c}{\makecell{Statistics}} \\
\cline{4-12}
 \multicolumn{1}{c}{\makecell{}} & \multicolumn{1}{c}{\makecell{}} &
& \makecell{Trend} 
& \makecell{Transition} 
& \makecell{Fluctuation} 
& \makecell{Outliers} 
& \makecell{Amplitude \\ level} 
& \makecell{Initial \\ value} 
& \makecell{Final \\ value} 
& \makecell{Temporal \\ mean} 
& \makecell{Temporal \\ standard \\ deviation} \\
\hline
\multicolumn{3}{c|}{\makecell{Single LIB cell}} 
& \makecell{\checkmark} 
& \makecell{\checkmark} 
& \makecell{\checkmark} 
& \makecell{\checkmark} 
& \makecell{} 
& \makecell{\checkmark} 
& \makecell{\checkmark} 
& \makecell{\checkmark} 
& \makecell{\checkmark} \\
\hline

\multirow{5}{*}{\makecell{System \\ level}} 
& \multirow{3}{*}{\makecell{Aggregate \\ behavior}} 
& \makecell{Mean} 
& \makecell{\checkmark} 
& \makecell{\checkmark} 
& \makecell{\checkmark} 
& \makecell{\checkmark} 
& \makecell{} 
& \makecell{\checkmark} 
& \makecell{\checkmark} 
& \makecell{\checkmark} 
& \makecell{\checkmark} \\
\cline{3-12}
& & \makecell{Maximum} 
& \makecell{\checkmark} 
& \makecell{\checkmark} 
& \makecell{} 
& \makecell{\checkmark} 
& \makecell{} 
& \makecell{\checkmark} 
& \makecell{\checkmark} 
& \makecell{\checkmark} 
& \makecell{\checkmark} \\
\cline{3-12}
& & \makecell{Minimum} 
& \makecell{\checkmark} 
& \makecell{\checkmark} 
& \makecell{} 
& \makecell{\checkmark} 
& \makecell{} 
& \makecell{\checkmark} 
& \makecell{\checkmark} 
& \makecell{\checkmark} 
& \makecell{\checkmark} \\
\cline{2-12}
& \multirow{2}{*}[-8pt]{\makecell{Heterogeneity}} 
& \makecell{Standard \\ deviation} 
& \makecell{} 
& \makecell{} 
& \makecell{\checkmark} 
& \makecell{} 
& \makecell{\checkmark} 
& \makecell{} 
& \makecell{} 
& \makecell{\checkmark} 
& \makecell{\checkmark} \\
\cline{3-12}
& & \makecell{Shannon \\ entropy} 
& \makecell{} 
& \makecell{} 
& \makecell{\checkmark} 
& \makecell{} 
& \makecell{\checkmark} 
& \makecell{} 
& \makecell{} 
& \makecell{\checkmark} 
& \makecell{\checkmark} \\
\hline
\end{tabular}
\end{adjustbox}\label{tab-attribute}
\end{table}

\subsection{Rules for assigning attribute descriptors}\label{assigning-rule}
To systematically assign semantic descriptors to each semantic attribute within a given temporal slice, we define a set of rule-based criteria, summarized in Table \ref{tab2}. These rules are grounded in established principles from standard time-series analysis literature \cite{best2007perception,miot2020empirical,borowski2014online,bertrand2011off,cousineau2010outliers,rousseeuw2011robust}, ensuring consistency and interpretability across time series. The principles for selecting hyperparameters are detailed in Supplementary Note 6 and Table S3.
\begin{table}[h!]
\centering
\caption{Rules for assigning attribute descriptors for time slices.}\label{tab2}\vspace*{1em}

\setlength{\tabcolsep}{6pt}
\renewcommand{\arraystretch}{1.25}

\begin{adjustbox}{max width=\textwidth,center}
\begin{tabular}{c| c >{\raggedright\arraybackslash}p{6.5cm} >{\centering\arraybackslash}p{3cm}}
\hline
\textbf{Attributes} & \textbf{Hyperparameters} & \multicolumn{1}{c}{\textbf{Assigning rules}} & \textbf{Descriptors} \\
\hline
\makecell{Trend} & \makecell{$\epsilon$} & 
\makecell[l]{Fit the time series using a linear function\\ $y=at+b$. If the slope $a$ exceeds a\\ positive threshold $\epsilon$, the trend is classified\\ as ``\textit{increasing}''; if $a<-\epsilon$, it is classified \\as ``\textit{decreasing}''. Otherwise, the trend is \\considered ``\textit{stable}''.} & 
\makecell{Increasing;\\ Decreasing;\\ Stable} \\
\hline
\makecell{Transition} & \makecell{$\omega$, $\xi$} & 
\makecell[l]{Estimate the local trend using a sliding\\ window of size $\omega$. Within each window, \\perform a first-order linear fit to compute\\ a local slope, resulting in a sequence of \\local slope estimates over time. A \\transition event is detected when the \\absolute difference between two successive\\ local slopes exceeds $\xi$. The temporal \\midpoint of the corresponding sliding \\window is defined as the transition time.} & 
\makecell{Transition at the\\$t^\text{th}$ time stamp;\\ Without \\transition} \\
\hline
Fluctuation & $\beta$ & 
\makecell[l]{Apply a linear fit to the data within each\\ time slice, modeling as $y=at+b$. \\Compute the detrended signal\\ $s'(t)=s(t)-at-b$. Calculate the\\ variance of $s'(t)$. If this variance exceeds\\ $\beta$, classify as noticeable fluctuation.} & 
\makecell{Fluctuation \\Remains slight;\\ With noticeable \\fluctuation} \\
\hline
Outlier
& $\vartheta$ 
& \makecell[l]{Outliers are detected by using the Z-score\\ method. The absolute deviation at each\\ time stamp is computed as $\Delta_i = |x_i - \bar{x}|$,\\ where $\bar{x} = \frac{1}{N}\sum_{i=1}^N x_i$. A time stamp is\\ identified as an outlier if its absolute if $\Delta_i$ \\exceeds both $3\sigma$ and a predefined \\threshold $\vartheta$, where \\$\sigma = \sqrt{\frac{1}{N-1}\sum_{i=1}^N (x_i - \bar{x})^2}$.}
& \makecell{Outlier detected at\\ the $t^\text{th}$ time stamp; \\Without outliers}\\
\hline
Amplitude level          & $\delta_1$, $\delta_2$ & \makecell[l]{Compute the absolute mean value of the \\signal and define two thresholds, $\delta_1$ and \\$\delta_2$ (where $\delta_2 > \delta_1 > 0$). If the absolute \\mean exceeds $\delta_2$, the signal is classified as \\``\textit{significant}''; if it lies between $\delta_1$ and $\delta_2$, \\it is classified as ``\textit{moderate}''; otherwise, it\\ is considered ``\textit{slight}''.} & \makecell{Significant;\\ Moderate; \\Slight} \\
\hline
Temporal mean            &    -                   & \makecell[l]{Calculate the average value of the time \\series across the duration of the time\\ slice.} & - \\
\hline
Temporal standard deviation &   -                  & \makecell[l]{Calculate the standard deviation of the \\time series across the duration of the time\\ slice.} & - \\
\hline
Initial value            &     -                  & \makecell[l]{Record the value of the time series at the \\start of the time slice.} & - \\
\hline
Final value              &     -                  & \makecell[l]{Record the value of the time series at the \\end of the time slice.} & - \\
\hline
\end{tabular}
\end{adjustbox}\label{tab-descriptor-assignment}
\end{table}

\subsection{Consolidation of redundant local descriptors}\label{consolidation-descriptor}
At this stage, each time series is represented as a sequence of semantic descriptors. To reduce redundancy while preserving descriptive quality, we perform a descriptor merging step that consolidates consecutive slices with identical semantics. Starting from the first time slice, we iteratively compare each descriptor with the next; if they are identical, the slices are merged into a single segment. For example, if the voltage of a LIB cell is consistently increasing across time slices 1 to 4, these slices are combined under the unified descriptor “\textit{increasing}”. 

\subsection{Translating descriptors into expressions}\label{translating-descriptors}
After merging, each descriptor is paired with its corresponding time slice and target time series to form a triplet:
\begin{itemize}
 \item Descriptor: Trend – increasing
 \item Target signal: Voltage of Cell $\#$1
 \item Time slice: $1^\text{st}$ to $50^\text{th}$ time stamps
\end{itemize}

This triplet is then converted into natural language using a predefined translation template (see Table S5). For the example above, the resulting expression is:

“\textit{From the $1^\text{st}$ to the $50^\text{th}$ time stamp, the voltage of Cell $\#$1 kept increasing.}”

\subsection{Converting expressions into reports}\label{convert-expression-into-report}

After the raw LIB time-series data are transformed into textual expressions in the previous stage, these expressions still appear as a sequence of record-style entries with limited readability and weak structural coherence. To resolve this issue, we further employ LLMs to reorganize these semantic expressions into structurally rigorous, fluent, and information-complete reports. Leveraging the intrinsic capability of LLMs for text understanding, synthesis, and summarization, this conversion can be accomplished autonomously without any retraining, which fundamentally distinguishes our framework from multimodal LLM approaches for time-series analysis that typically rely on task-specific fine-tuning.

The report generation process takes two inputs: (\romannumeral1) the textual expressions produced in the previous stage, and (\romannumeral2) an expert-guided instruction specifying the target reporting task (see Supplementary Note~7). These inputs are provided to the LLM to synthesize descriptive, multi-attribute reports corresponding to the original time-series data. This procedure is model-agnostic and can be readily implemented with any mainstream LLM through a simple API call.

To improve computational efficiency, we design the API interface to support parallel report generation across multiple single cells. Specifically, for the MIT and TJU datasets, we consider reporting tasks for 6 and 4 standalone LIB cells, respectively, and the time-series data from different cells are jointly provided as inputs, prompting the LLM to generate cell-wise reports within a single API call. For the ZJU-t dataset, a similar parallel strategy is adopted. However, due to the maximum token limitation of current LLMs, multiple independent API calls are issued to separately generate reports for groups of cells: (1) cells \#1--4, (2) cells \#5--8, (3) cells \#9--12, and (4) cells \#13--16. In addition, a system-level time-series report is generated through an independent API call. All report segments, including the four single-cell reports and the system-level report, are finally concatenated to form a complete report for the entire LIB system.

\subsection{Time-series reports for downstream battery O\&M applications}\label{ref4-8}

The generated time-series reports constitute a rich and structured textual representation of LIB operational behaviours, which can be directly integrated into mainstream LLMs for explainable battery O\&M applications. Specifically, we supply the LLM with (\romannumeral1) task-specific human queries (see Supplementary Note~5) and (\romannumeral2) the corresponding report text as prompts. Within a question--answering framework, the LLM processes these inputs to produce natural-language O\&M responses.

For more specialized tasks such as fault diagnosis, expert knowledge can be further incorporated through advanced prompting strategies or RAG. This framework enables a unified, interpretable, and knowledge-enhanced interface between raw time-series data and practical battery O\&M decision-making.

\section*{Conflict of interests}
The authors declare that they have no conflict of interest.

\section*{Acknowledgements}
J.Y. and C.Z. acknowledge support from the National Natural Science Foundation of China (Grant Nos. 62125306, 62450020, and U25B2050) and the State Key Laboratory of Industrial Control Technology, China (Grant No. ICT2025C01). Z.C. acknowledges support from a Natural Science and Engineering Research Council of Canada Discovery Grant (RGPIN-2024-06015). All authors gratefully acknowledge the members of Prof. Zhao’s group at ZJU and Prof. Furong Gao’s group at Hong Kong University of Science and Technology (HKUST) for their assistance with time-series data annotation.

\section*{Author contributions}

Conceptualisation, J.Yang and C.Zhao; Methodology, J.Yang and C.Zhao; Software, J.Yang and C.Zhao; Original data acquisition, J.Yang and C.Zhao; Analysis, J.Yang, C.Zhao, M.Guay and Z.Cao; Visualisation, J.Yang and C.Zhao; Writing original draft, J.Yang, C.Zhao, M.Guay and Z.Cao; Writing, review \& editing, J.Yang, C.Zhao, M.Guay and Z.Cao; Supervision, C.Zhao and Z.Cao.

\section*{References}
\printbibliography[heading=none]{}

@article{harper2019recycling,
  title={Recycling lithium-ion batteries from electric vehicles},
  author={Harper, Gavin and Sommerville, Roberto and Kendrick, Emma and Driscoll, Laura and Slater, Peter and Stolkin, Rustam and Walton, Allan and Christensen, Paul and Heidrich, Oliver and Lambert, Simon and others},
  journal={Nature},
  volume={575},
  number={7781},
  pages={75--86},
  year={2019},
  publisher={Nature Publishing Group}
}

@article{liang2019review,
  title={A review of rechargeable batteries for portable electronic devices},
  author={Liang, Yeru and Zhao, Chen-Zi and Yuan, Hong and Chen, Yuan and Zhang, Weicai and Huang, Jia-Qi and Yu, Dingshan and Liu, Yingliang and Titirici, Maria-Magdalena and Chueh, Yu-Lun and others},
  journal={InfoMat},
  volume={1},
  number={1},
  pages={6--32},
  year={2019},
  publisher={Wiley Online Library}
}

@article{arteaga2017overview,
  title={Overview of lithium-ion grid-scale energy storage systems},
  author={Arteaga, Juan and Zareipour, Hamidreza and Thangadurai, Venkataraman},
  journal={Current Sustainable/Renewable Energy Reports},
  volume={4},
  number={4},
  pages={197--208},
  year={2017},
  publisher={Springer}
}

@article{jiang2025battery,
  title={Battery technologies for grid-scale energy storage},
  author={Jiang, Taoli and Shen, Dongyang and Zhang, Zuodong and Liu, Hongxu and Zhao, Guili and Wang, Yidi and Tan, Shunxin and Luo, Ruihao and Chen, Wei},
  journal={Nature Reviews Clean Technology},
  pages={1--19},
  year={2025},
  publisher={Nature Publishing Group UK London}
}

@article{hu2020battery,
  title={Battery lifetime prognostics},
  author={Hu, Xiaosong and Xu, Le and Lin, Xianke and Pecht, Michael},
  journal={Joule},
  volume={4},
  number={2},
  pages={310--346},
  year={2020},
  publisher={Elsevier}
}

@article{severson2019data,
  title={Data-driven prediction of battery cycle life before capacity degradation},
  author={Severson, Kristen A and Attia, Peter M and Jin, Norman and Perkins, Nicholas and Jiang, Benben and Yang, Zi and Chen, Michael H and Aykol, Muratahan and Herring, Patrick K and Fraggedakis, Dimitrios and others},
  journal={Nature Energy},
  volume={4},
  number={5},
  pages={383--391},
  year={2019},
  publisher={Nature Publishing Group UK London}
}

@article{gu2023novel,
  title={A novel state-of-health estimation for the lithium-ion battery using a convolutional neural network and transformer model},
  author={Gu, Xinyu and See, Khay Wai and Li, Penghua and Shan, Kangheng and Wang, Yunpeng and Zhao, Liang and Lim, Kai Chin and Zhang, Neng},
  journal={Energy},
  volume={262},
  pages={125501},
  year={2023},
  publisher={Elsevier}
}

@article{che2022state,
  title={State of health prognostics for series battery packs: A universal deep learning method},
  author={Che, Yunhong and Deng, Zhongwei and Li, Penghua and Tang, Xiaolin and Khosravinia, Kavian and Lin, Xianke and Hu, Xiaosong},
  journal={Energy},
  volume={238},
  pages={121857},
  year={2022},
  publisher={Elsevier}
}

@article{cao2025model,
  title={Model-constrained deep learning for online fault diagnosis in {Li}-ion batteries over stochastic conditions},
  author={Cao, Rui and Zhang, Zhengjie and Shi, Runwu and Lu, Jiayi and Zheng, Yifan and Sun, Yefan and Liu, Xinhua and Yang, Shichun},
  journal={Nature Communications},
  volume={16},
  number={1},
  pages={1651},
  year={2025},
  publisher={Nature Publishing Group UK London}
}

@article{zhang2023realistic,
  title={Realistic fault detection of li-ion battery via dynamical deep learning},
  author={Zhang, Jingzhao and Wang, Yanan and Jiang, Benben and He, Haowei and Huang, Shaobo and Wang, Chen and Zhang, Yang and Han, Xuebing and Guo, Dongxu and He, Guannan and others},
  journal={Nature Communications},
  volume={14},
  number={1},
  pages={5940},
  year={2023},
  publisher={Nature Publishing Group UK London}
}

@article{yang2025toward,
  title={Toward the ensemble consistency: Condition-driven ensemble balance representation learning and nonstationary anomaly detection for battery energy storage system},
  author={Yang, Jiayang and Chen, Xu and Zhao, Chunhui},
  journal={Applied Energy},
  volume={381},
  pages={125160},
  year={2025},
  publisher={Elsevier}
}

@article{xie2022faults,
  title={Faults diagnosis for large-scale battery packs via texture analysis on spatial--temporal images converted from electrical behaviors},
  author={Xie, Jiale and Wang, Guang and Liu, Jun and Li, Zengchao and Wei, Zhongbao},
  journal={IEEE Transactions on Transportation Electrification},
  volume={9},
  number={4},
  pages={4876--4887},
  year={2022},
  publisher={IEEE}
}

@article{wang2025data,
  title={Data-driven energy management for electric vehicles using offline reinforcement learning},
  author={Wang, Yong and Wu, Jingda and He, Hongwen and Wei, Zhongbao and Sun, Fengchun},
  journal={Nature Communications},
  volume={16},
  number={1},
  pages={2835},
  year={2025},
  publisher={Nature Publishing Group UK London}
}

@article{park2022deep,
  title={A deep reinforcement learning framework for fast charging of Li-ion batteries},
  author={Park, Saehong and Pozzi, Andrea and Whitmeyer, Michael and Perez, Hector and Kandel, Aaron and Kim, Geumbee and Choi, Yohwan and Joe, Won Tae and Raimondo, Davide M and Moura, Scott},
  journal={IEEE Transactions on Transportation Electrification},
  volume={8},
  number={2},
  pages={2770--2784},
  year={2022},
  publisher={IEEE}
}

@article{hao2023adaptive,
  title={Adaptive model-based reinforcement learning for fast-charging optimization of lithium-ion batteries},
  author={Hao, Yuhan and Lu, Qiugang and Wang, Xizhe and Jiang, Benben},
  journal={IEEE Transactions on Industrial Informatics},
  volume={20},
  number={1},
  pages={127--137},
  year={2023},
  publisher={IEEE}
}

@article{tan2019transfer,
  title={Transfer learning with long short-term memory network for state-of-health prediction of lithium-ion batteries},
  author={Tan, Yandan and Zhao, Guangcai},
  journal={IEEE Transactions on Industrial Electronics},
  volume={67},
  number={10},
  pages={8723--8731},
  year={2019},
  publisher={IEEE}
}

@article{mazzi2024lithium,
  title={Lithium-ion battery state of health estimation using a hybrid model based on a convolutional neural network and bidirectional gated recurrent unit},
  author={Mazzi, Yahia and Sassi, Hicham Ben and Errahimi, Fatima},
  journal={Engineering Applications of Artificial Intelligence},
  volume={127},
  pages={107199},
  year={2024},
  publisher={Elsevier}
}

@article{wang2024physics,
  title={Physics-informed neural network for lithium-ion battery degradation stable modeling and prognosis},
  author={Wang, Fujin and Zhai, Zhi and Zhao, Zhibin and Di, Yi and Chen, Xuefeng},
  journal={Nature Communications},
  volume={15},
  number={1},
  pages={4332},
  year={2024},
  publisher={Nature Publishing Group UK London}
}

@article{nascimento2021hybrid,
  title={Hybrid physics-informed neural networks for lithium-ion battery modeling and prognosis},
  author={Nascimento, Renato G and Corbetta, Matteo and Kulkarni, Chetan S and Viana, Felipe AC},
  journal={Journal of Power Sources},
  volume={513},
  pages={230526},
  year={2021},
  publisher={Elsevier}
}

@article{jiang2022fault,
  title={Fault diagnosis method for lithium-ion batteries in electric vehicles based on isolated forest algorithm},
  author={Jiang, Jiuchun and Li, Taiyu and Chang, Chun and Yang, Chen and Liao, Li},
  journal={Journal of Energy Storage},
  volume={50},
  pages={104177},
  year={2022},
  publisher={Elsevier}
}

@article{lee2022state,
  title={State-of-health estimation of Li-ion batteries in the early phases of qualification tests: An interpretable machine learning approach},
  author={Lee, Gyumin and Kim, Juram and Lee, Changyong},
  journal={Expert Systems with Applications},
  volume={197},
  pages={116817},
  year={2022},
  publisher={Elsevier}
}

@article{zhao2025lithium,
  title={Lithium-ion battery remaining useful life prediction based on interpretable deep learning and network parameter optimization},
  author={Zhao, Bo and Zhang, Weige and Zhang, Yanru and Zhang, Caiping and Zhang, Chi and Zhang, Junwei},
  journal={Applied Energy},
  volume={379},
  pages={124713},
  year={2025},
  publisher={Elsevier}
}

@article{wang2023explainability,
  title={Explainability-driven model improvement for SOH estimation of lithium-ion battery},
  author={Wang, Fujin and Zhao, Zhibin and Zhai, Zhi and Shang, Zuogang and Yan, Ruqiang and Chen, Xuefeng},
  journal={Reliability Engineering \& System Safety},
  volume={232},
  pages={109046},
  year={2023},
  publisher={Elsevier}
}

@article{naveed2025comprehensive,
  title={A comprehensive overview of large language models},
  author={Naveed, Humza and Khan, Asad Ullah and Qiu, Shi and Saqib, Muhammad and Anwar, Saeed and Usman, Muhammad and Akhtar, Naveed and Barnes, Nick and Mian, Ajmal},
  journal={ACM Transactions on Intelligent Systems and Technology},
  volume={16},
  number={5},
  pages={1--72},
  year={2025},
  publisher={ACM New York, NY}
}

@article{li2025self,
  title={Self-reflection enhances large language models towards substantial academic response},
  author={Li, Baoxue and Zhao, Chunhui},
  journal={npj Artificial Intelligence},
  volume={1},
  number={1},
  pages={42},
  year={2025},
  publisher={Nature Publishing Group UK London}
}

@article{guo2025deepseek,
  title={Deepseek-{R}1 incentivizes reasoning in {LLM}s through reinforcement learning},
  author={Guo, Daya and Yang, Dejian and Zhang, Haowei and Song, Junxiao and Wang, Peiyi and Zhu, Qihao and Xu, Runxin and Zhang, Ruoyu and Ma, Shirong and Bi, Xiao and others},
  journal={Nature},
  volume={645},
  number={8081},
  pages={633--638},
  year={2025},
  publisher={Nature Publishing Group UK London}
}

@article{achiam2023gpt,
  title={{GPT}-4 technical report},
  author={Achiam, Josh and Adler, Steven and Agarwal, Sandhini and Ahmad, Lama and Akkaya, Ilge and Aleman, Florencia Leoni and Almeida, Diogo and Altenschmidt, Janko and Altman, Sam and Anadkat, Shyamal and others},
  journal={arXiv preprint arXiv:2303.08774},
  year={2023}
}

@inproceedings{jiang2024empowering,
  title={Empowering time series analysis with large language models: A survey},
  author={Jiang, Yushan and Pan, Zijie and Zhang, Xikun and Garg, Sahil and Schneider, Anderson and Nevmyvaka, Yuriy and Song, Dongjin},
  booktitle={Proceedings of the Thirty-Third International Joint Conference on Artificial Intelligence},
  pages={8095--8103},
  year={2024}
}

@article{wang2025itformer,
  title={{ITFormer}: Bridging Time Series and Natural Language for Multi-Modal {QA} with Large-Scale Multitask Dataset},
  author={Wang, Yilin and Lei, Peixuan and Song, Jie and Hao, Yuzhe and Chen, Tao and Zhang, Yuxuan and Jia, Lei and Li, Yuanxiang and Wei, Zhongyu},
  journal={arXiv preprint arXiv:2506.20093},
  year={2025}
}

@article{jin2023time,
  title={Time-{LLM}: Time series forecasting by reprogramming large language models},
  author={Jin, Ming and Wang, Shiyu and Ma, Lintao and Chu, Zhixuan and Zhang, James Y and Shi, Xiaoming and Chen, Pin-Yu and Liang, Yuxuan and Li, Yuan-Fang and Pan, Shirui and others},
  journal={arXiv preprint arXiv:2310.01728},
  year={2023}
}

@article{xie2024chatts,
  title={Chat{TS}: Aligning time series with {LLM}s via synthetic data for enhanced understanding and reasoning},
  author={Xie, Zhe and Li, Zeyan and He, Xiao and Xu, Longlong and Wen, Xidao and Zhang, Tieying and Chen, Jianjun and Shi, Rui and Pei, Dan},
  journal={arXiv preprint arXiv:2412.03104},
  year={2024}
}

@article{shen2017co,
  title={The co-estimation of state of charge, state of health, and state of function for lithium-ion batteries in electric vehicles},
  author={Shen, Ping and Ouyang, Minggao and Lu, Languang and Li, Jianqiu and Feng, Xuning},
  journal={IEEE Transactions on Vehicular Technology},
  volume={67},
  number={1},
  pages={92--103},
  year={2017},
  publisher={IEEE}
}

@article{adaikkappan2022modeling,
  title={Modeling, state of charge estimation, and charging of lithium-ion battery in electric vehicle: A review},
  author={Adaikkappan, Maheshwari and Sathiyamoorthy, Nageswari},
  journal={International Journal of Energy Research},
  volume={46},
  number={3},
  pages={2141--2165},
  year={2022},
  publisher={Wiley Online Library}
}

@article{mei2023operando,
  title={Operando monitoring of thermal runaway in commercial lithium-ion cells via advanced lab-on-fiber technologies},
  author={Mei, Wenxin and Liu, Zhi and Wang, Chengdong and Wu, Chuang and Liu, Yubin and Liu, Pengjie and Xia, Xudong and Xue, Xiaobin and Han, Xile and Sun, Jinhua and others},
  journal={Nature Communications},
  volume={14},
  number={1},
  pages={5251},
  year={2023},
  publisher={Nature Publishing Group UK London}
}

@article{mama2025comprehensive,
  title={Comprehensive review of multi-scale Lithium-ion batteries modeling: From electro-chemical dynamics up to heat transfer in battery thermal management system},
  author={Mama, Magui and Solai, Elie and Capurso, Tommaso and Danlos, Amelie and Khelladi, Sofiane},
  journal={Energy Conversion and Management},
  volume={325},
  pages={119223},
  year={2025},
  publisher={Elsevier}
}

@inproceedings{kil1997optimum,
  title={Optimum window size for time series prediction},
  author={Kil, Rhee M and Park, Seon Hee and Kim, Seunghwan},
  booktitle={Proceedings of the 19th Annual International Conference of the IEEE Engineering in Medicine and Biology Society.'Magnificent Milestones and Emerging Opportunities in Medical Engineering'(Cat. No. 97CH36136)},
  volume={4},
  pages={1421--1424},
  year={1997},
  organization={IEEE}
}

@article{zhu2022data,
  title={Data-driven capacity estimation of commercial lithium-ion batteries from voltage relaxation},
  author={Zhu, Jiangong and Wang, Yixiu and Huang, Yuan and Bhushan Gopaluni, R and Cao, Yankai and Heere, Michael and M{\"u}hlbauer, Martin J and Mereacre, Liuda and Dai, Haifeng and Liu, Xinhua and others},
  journal={Nature Communications},
  volume={13},
  number={1},
  pages={2261},
  year={2022},
  publisher={Nature Publishing Group UK London}
}

@inproceedings{min2023factscore,
  title={{FactScore}: Fine-grained atomic evaluation of factual precision in long form text generation},
  author={Min, Sewon and Krishna, Kalpesh and Lyu, Xinxi and Lewis, Mike and Yih, Wen-tau and Koh, Pang and Iyyer, Mohit and Zettlemoyer, Luke and Hajishirzi, Hannaneh},
  booktitle={Proceedings of the 2023 Conference on Empirical Methods in Natural Language Processing},
  pages={12076--12100},
  year={2023}
}

@article{dubey2024llama,
  title={The {Llama} 3 herd of models},
  author={Dubey, Abhimanyu and Jauhri, Abhinav and Pandey, Abhinav and Kadian, Abhishek and Al-Dahle, Ahmad and Letman, Aiesha and Mathur, Akhil and Schelten, Alan and Yang, Amy and Fan, Angela and others},
  journal={arXiv e-prints},
  pages={arXiv--2407},
  year={2024}
}

@article{yang2025qwen3,
  title={Qwen3 technical report},
  author={Yang, An and Li, Anfeng and Yang, Baosong and Zhang, Beichen and Hui, Binyuan and Zheng, Bo and Yu, Bowen and Gao, Chang and Huang, Chengen and Lv, Chenxu and others},
  journal={arXiv preprint arXiv:2505.09388},
  year={2025}
}

@article{zeng2025glm,
  title={{GLM}-4.5: Agentic, reasoning, and coding (arc) foundation models},
  author={Zeng, Aohan and Lv, Xin and Zheng, Qinkai and Hou, Zhenyu and Chen, Bin and Xie, Chengxing and Wang, Cunxiang and Yin, Da and Zeng, Hao and Zhang, Jiajie and others},
  journal={arXiv preprint arXiv:2508.06471},
  year={2025}
}

@article{miot2020empirical,
  title={An empirical study of neural networks for trend detection in time series},
  author={Miot, Alexandre and Drigout, Gilles},
  journal={SN Computer Science},
  volume={1},
  number={6},
  pages={347},
  year={2020},
  publisher={Springer}
}

@article{bertrand2011off,
  title={Off-line detection of multiple change points by the filtered derivative with p-value method},
  author={Bertrand, Pierre Raphael and Fhima, Mehdi and Guillin, Arnaud},
  journal={Sequential Analysis},
  volume={30},
  number={2},
  pages={172--207},
  year={2011},
  publisher={Taylor \& Francis}
}

@article{borowski2014online,
  title={Online signal extraction by robust regression in moving windows with data-adaptive width selection: {SCARM}—{S}lope Comparing Adaptive Repeated Median},
  author={Borowski, Matthias and Fried, Roland},
  journal={Statistics and Computing},
  volume={24},
  number={4},
  pages={597--613},
  year={2014},
  publisher={Springer}
}

@article{cousineau2010outliers,
  title={Outliers detection and treatment: A review.},
  author={Cousineau, Denis and Chartier, Sylvain},
  journal={International Journal of Psychological Research},
  volume={3},
  number={1},
  pages={58--67},
  year={2010},
  publisher={Universidad de San Buenaventura}
}

@article{best2007perception,
  title={Perception of linear and nonlinear trends: Using slope and curvature information to make trend discriminations},
  author={Best, Lisa A and Smith, Laurence D and Stubbs, D Alan},
  journal={Perceptual and Motor Skills},
  volume={104},
  number={3},
  pages={707--721},
  year={2007},
  publisher={SAGE Publications Sage CA: Los Angeles, CA}
}

@article{rousseeuw2011robust,
  title={Robust statistics for outlier detection},
  author={Rousseeuw, Peter J and Hubert, Mia},
  journal={Wiley Interdisciplinary Reviews: Data Mining and Knowledge Discovery},
  volume={1},
  number={1},
  pages={73--79},
  year={2011},
  publisher={Wiley Online Library}
}

@article{hong2025glm,
  title={{GLM}-4.1{V}-{Thinking}: Towards versatile multimodal reasoning with scalable reinforcement learning},
  author={Hong, Wenyi and Yu, Wenmeng and Gu, Xiaotao and Wang, Guo and Gan, Guobing and Tang, Haomiao and Cheng, Jiale and Qi, Ji and Ji, Junhui and Pan, Lihang and others},
  journal={arXiv preprint arXiv:2507.01006},
  year={2025}
}

@article{ng2009enhanced,
  title={Enhanced coulomb counting method for estimating state-of-charge and state-of-health of lithium-ion batteries},
  author={Ng, Kong Soon and Moo, Chin-Sien and Chen, Yi-Ping and Hsieh, Yao-Ching},
  journal={Applied Energy},
  volume={86},
  number={9},
  pages={1506--1511},
  year={2009},
  publisher={Elsevier}
}

@INPROCEEDINGS{11268252,
  author={Li, Baoxue and Zhao, Chunhui},
  booktitle={2025 CAA Symposium on Fault Detection, Supervision, and Safety for Technical Processes (SAFEPROCESS)}, 
  title={{S2S-FDD}: Bridging Industrial Time Series and Natural Language for Explainable Zero-shot Fault Diagnosis}, 
  year={2025},
  volume={},
  number={},
  pages={1-6},
  doi={10.1109/SAFEPROCESS67117.2025.11268252}}

@article{zhao2025align,
  title={Align knowledge with time-series: Cross-modal domain knowledge activation for {LLM}-enabled zero-shot fault diagnosis},
  author={Zhao, Jiancheng and Zhao, Chunhui and Yue, Jiaqi},
  journal={Journal of Process Control},
  volume={155},
  pages={103534},
  year={2025},
  publisher={Elsevier}
}

@article{chen2019recycling,
  title={Recycling end-of-life electric vehicle lithium-ion batteries},
  author={Chen, Mengyuan and Ma, Xiaotu and Chen, Bin and Arsenault, Renata and Karlson, Peter and Simon, Nakia and Wang, Yan},
  journal={Joule},
  volume={3},
  number={11},
  pages={2622--2646},
  year={2019},
  publisher={Elsevier}
}

@article{li2019data,
  title={Data-Driven Safety Envelope of Lithium-Ion Batteries for Electric Vehicles},
  author={Li, Wei and Zhu, Juner and Xia, Yong and B. Gorji, Maysam and Wierzbicki, Tomasz},
  journal={Joule},
  volume={3},
  number={11},
  pages={2703--2715},
  year={2019},
  publisher={Elsevier}
}

@article{rhyu2025systematic,
  title={Systematic feature design for cycle life prediction of lithium-ion batteries during formation},
  author={Rhyu, Jinwook and Schaeffer, Joachim and Li, Michael L and Cui, Xiao and Chueh, William C and Bazant, Martin Z and Braatz, Richard D},
  journal={Joule},
  volume={9},
  number={5},
  year={2025},
  publisher={Elsevier}
}

@article{majumder2024exploring,
  title={Exploring the capabilities and limitations of large language models in the electric energy sector},
  author={Majumder, Subir and Dong, Lin and Doudi, Fatemeh and Cai, Yuting and Tian, Chao and Kalathil, Dileep and Ding, Kevin and Thatte, Anupam A and Li, Na and Xie, Le},
  journal={Joule},
  volume={8},
  number={6},
  pages={1544--1549},
  year={2024},
  publisher={Elsevier}
}

@article{zuo2025large,
  title={Large language models for batteries},
  author={Zuo, Wenhua and Zheng, Huihuo and He, Tanjin and Vishwanath, Venkatram and Chan, Maria KY and Stevens, Rick L and Amine, Khalil and Xu, Gui-Liang},
  journal={Joule},
  volume={9},
  number={8},
  year={2025},
  publisher={Elsevier}
}

\end{document}


\linenumbers

\maketitle
\setcounter{tocdepth}{1}
\tableofcontents
\newpage

\renewcommand{\thefigure}{S\arabic{figure}}
\setcounter{figure}{0}

\renewcommand{\thetable}{S\arabic{table}}
\setcounter{table}{0}

\section{Supplementary Note 1. Data curation and preprocessing description for MIT and TJU datasets}\label{supplementary_note_1}


We sourced two publicly available lithium-ion battery (LIB) datasets from Refs.~\cite{severson2019data, zhu2022data}. In Ref.~\cite{severson2019data}, data were collected from 124 standalone lithium iron phosphate (LFP) commercial LIB cells, each cycled to failure using a 48-channel Arbin LBT potentiostat housed in a forced-convection chamber maintained at 30\textdegree{}C. From this collection, six cells (channels 1--3 from the 2017-05-12 batch and channels 1--3 from the 2017-06-30 batch) were selected as representative samples and processed into what we refer to as the \textit{MIT dataset}.

In Ref.~\cite{zhu2022data}, operational data were obtained from 130 standalone LIB cells, including 66 nickel cobalt aluminum oxide (NCA)-type, 55 nickel cobalt manganese oxide (NCM)-type, and 9 NCA+NCM-type variants. These cells were cycled under temperature-controlled conditions. For this study, we selected four NCA-type cells (tags: CY45-05\_1-\#1, \#2, \#3, \#4) as representative samples and processed them into the \textit{TJU dataset}. Metadata for both datasets are summarized in Table~\ref{si table: MIT metadata}.

\begin{table}[h]
\centering
\caption{Summary of metadata for the MIT and TJU LIB datasets.}\label{si table: MIT metadata}\vspace*{1em}

\setlength{\tabcolsep}{6pt}
\renewcommand{\arraystretch}{1.25}

\begin{adjustbox}{max width=\textwidth,center}
\begin{tabular}{c c c }
\toprule
Property&MIT dataset&TJU dataset\\
\toprule
Chemical component&LiFePO$_4$&LiNiCoAlO$_2$\\
Nominal capacity (Ah)&1.1&3.5\\
Cut-off voltage (V)&2.0-3.6&2.65-4.2\\
Experiment temperature (°C)&30&45\\
Charging/discharging protocol&CC-CV/CC&CC-CV/CC\\
Number of cells & 6 & 4\\
Number of cycles per cell & 10 & 10\\
Sampling period & 5 seconds & 10 seconds\\
Section duration & 500 seconds & 1000 seconds\\
\bottomrule
\end{tabular}
\end{adjustbox}
\end{table}

\begin{figure}[h!]
\centering
\includegraphics[width=0.6\textwidth]{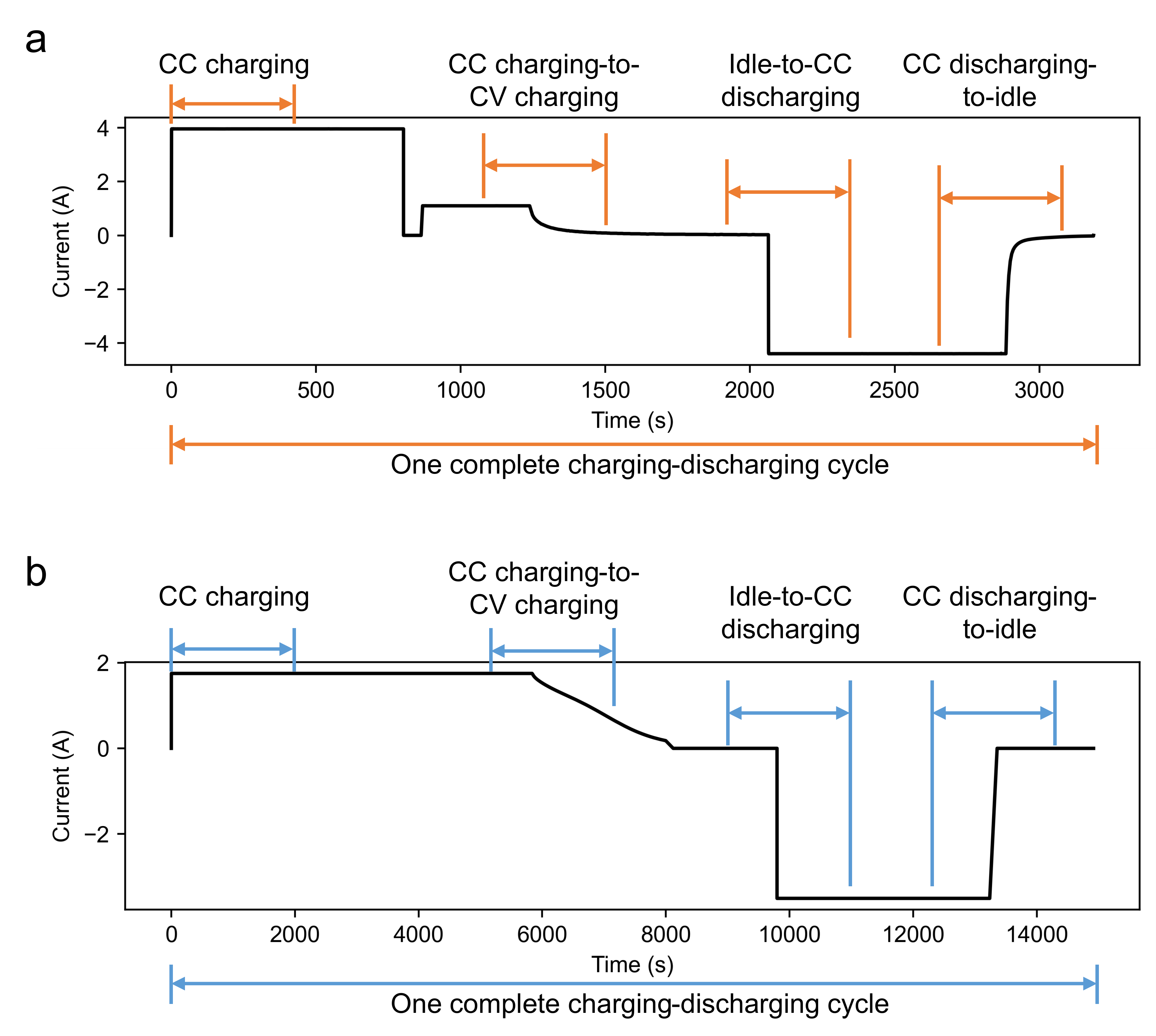}
\caption{Representative cycle segments from the (a) MIT and (b) TJU datasets.
Illustration of four characteristic operating sections extracted from a typical charge-discharge cycle for each dataset: (\romannumeral1) constant-current (CC) charging, (\romannumeral2) CC-to-constant-voltage (CV) transition, (\romannumeral3) idle-to-CC discharging transition, and (\romannumeral4) CC discharging-to-idle transition. Each section spans 100 time steps, with a sampling interval of 5 seconds for the MIT dataset and 10 seconds for the TJU dataset. These segments reflect the dynamic behavior under different operational modes and form the basis for evaluating the performance of TimeSeries2Report.}\label{figs1}
\end{figure}

All selected cells across both datasets were subjected to an identical charge/discharge protocol: constant-current constant-voltage (CC-CV) charging followed by constant-current (CC) discharging. This results in operational cycles exhibiting consistent dynamic patterns. To reduce the annotation workload for human experts, we randomly selected 10 complete charge-discharge cycles per representative cell. Within each cycle, we extracted four characteristic segments corresponding to distinct operating conditions: (\romannumeral1) CC charging, (\romannumeral2) CC-to-CV charging transition, (\romannumeral3) idle-to-CC discharging transition, and (\romannumeral4) CC discharging-to-idle transition. Each segment consists of 100 time steps, with sampling intervals of 5~seconds for the MIT dataset and 10~seconds for the TJU dataset. Representative examples of these segments are illustrated in Fig.~\ref{figs1}. This completes the construction of the curated MIT and TJU datasets.

\clearpage

\section{Supplementary Note 2. Ground-truth annotation pipeline by human experts}
\label{supplementary_note_2}

\subsection{Expert annotation pipeline for battery time-series reports}

To evaluate the quality of TS2R-generated reports, we engaged domain experts to independently generate human-written reports from raw battery time-series data. These expert-generated reports serve as high-confidence ground truth references for factual consistency evaluation. The overall annotation pipeline is described below.

We implemented a structured two-stage annotation process involving two independent expert groups: two second-year PhD students from Prof. Furong Gao's group at HKUST, and three second-year PhD students from Prof. Chunhui Zhao's group at ZJU – all with domain expertise in LIB operation and diagnostics. An initial consensus meeting was held to define the annotation standards and establish consistent definitions for temporal descriptors. The annotation process then proceeded in two stages: Group 1 (three experts) conducted the initial independent annotations, while Group 2 (two experts) performed quality control and synthesized the final reference reports.

\subsubsection{Stage 1: Initial annotation writing}

Group 1 experts were provided with raw time-series data, including both numerical values and visual plots, along with a standardized annotation questionnaire. Each time-series variable was annotated across seven key semantic attributes: \textit{Trend}, \textit{Transition Events}, \textit{Fluctuation}, \textit{Outliers}, \textit{Amplitude Level}, \textit{Initial/Final Values}, and \textit{Temporal Statistics} (mean and standard deviation).

The formal procedures were as follows:

\begin{itemize}
    \item \textbf{Trend segmentation and characterization}: Experts segmented each time series into intervals of distinct temporal trends: 
    (\romannumeral1) Increasing, 
    (\romannumeral2) Decreasing, and 
    (\romannumeral3) Stable. 
    Quantitative slope thresholds (Table~2) guided segmentation consistency. For each segment, experts documented the trend type and exact time range in a standardized format:\\
    \centerline{\textit{``Trend: increase (1–50th time), stable (51–100th time).''}}

    \item \textbf{Transition event identification}: Transitions were marked at the boundaries between adjacent trend segments. Experts also identified multi-stage transitions within a segment if the rate of change varied significantly. These were recorded as:\\
    \centerline{\textit{``Transitions: A transition is observed at the 5th time.''}}

    \item \textbf{Fluctuation level assessment}: Using a sliding window (width = 10 time stamps), experts calculated the standard deviation within each window. A window was marked as fluctuating if it exceeded a predefined threshold. For example:\\
    \centerline{\textit{``Fluctuations: Noticeable fluctuations from 1–10th time.''}}\\
    If no significant fluctuations were found:\\
    \centerline{\textit{``Fluctuations: Fluctuations remain slight over the entire period.''}}

    \item \textbf{Outlier detection}: Experts first visually identified candidate outliers, then validated them statistically. A point was confirmed as an outlier if its local z-score (within a centered window of 11 time stamps) exceeded 3. Annotated as:\\
    \centerline{\textit{``Outliers: An outlier is observed at the 5th time.''}}\\
    Otherwise:\\
    \centerline{\textit{``Outliers: No outliers were detected.''}}

    \item \textbf{Amplitude level classification}: Applied to time series representing system-level heterogeneity (e.g., standard deviation, Shannon entropy across cells). Experts divided the series into 10-timestep windows and calculated the absolute mean of each slice. Based on thresholds in Table~2, amplitude was classified as slight, moderate, or significant. Example:\\
    \centerline{\textit{``Amplitude level: slight (1–50th time), moderate (51–100th time).''}}

    \item \textbf{Reporting of statistical descriptors}: For increasing/decreasing segments, initial and final values were recorded:\\
    \centerline{\textit{``Initial/final values: 3.3629 V to 3.4046 V (1–60th time).''}}\\
    For stable segments, the temporal mean and standard deviation were provided:\\
    \centerline{\textit{``Temporal mean and std.: Stable at 3.3387 V, std. is 0.0007 (71–100th time).''}}
\end{itemize}

\subsubsection{Stage 2: Quality control and report finalization}

The three Group 1 reports for each time series were consolidated into a unified questionnaire. Group 2 experts received these alongside the original time-series data and were tasked with selecting the most accurate description for each variable and attribute.

\paragraph{Independent parallel assessment:}
Each Group 2 expert independently reviewed the Group 1 reports and evaluated the factual accuracy and quantitative precision of each annotation. For each attribute, they selected the most accurate version, which could differ across attributes and across original annotators.

\paragraph{Consensus meeting and final synthesis:}
The two experts then met to reconcile discrepancies. If both selected the same annotation for an attribute, it was adopted. If disagreement arose, they re-examined the data using the criteria in Table~2 to reach consensus. If none of the original annotations were satisfactory, they performed a joint de novo analysis to generate a corrected version.

This process ensured high accuracy, internal consistency, and expert consensus in the final ground-truth report.

\subsection{Operational status labeling for anomaly detection ground truth}

To evaluate the TS2R framework in anomaly detection tasks, the same five experts also performed structured labeling of battery operational status.

\subsubsection{Expert annotation procedure}

Each expert independently assessed all time-series instances in the ZJU test dataset. The task involved two steps:

\begin{itemize}
    \item \textbf{Binary operational status labeling:} Experts assigned each instance a binary label (\texttt{Normal} or \texttt{Abnormal}) based on a holistic evaluation of the multivariate time series, informed by domain knowledge of standard LIB behavior.

    \item \textbf{Symptom description (for abnormal cases):} For instances labeled \texttt{Abnormal}, experts briefly described the anomaly, identifying the affected variable(s), cell ID(s), and providing a short label (e.g., ``thermal runaway onset'').
\end{itemize}

Annotations were collected using a standardized digital form to ensure reporting consistency.

\subsubsection{Ground truth consolidation via majority voting}

After annotation, the final operational status labels were determined by majority vote:

\begin{itemize}
    \item Instances labeled \texttt{Normal} by three or more experts were assigned a final label of \texttt{Normal}.
    \item Instances labeled \texttt{Abnormal} by three or more experts were assigned a final label of \texttt{Abnormal}.
\end{itemize}

For each \texttt{Abnormal} instance, the symptom descriptions from the majority group were aggregated to create a comprehensive fault profile. This adjudicated dataset serves as the high-fidelity ground truth for benchmarking TS2R-based anomaly detection performance.

\newpage
\section{Supplementary Note 3. Prompts for quantitative evaluation of generated report}\label{supplementary_note_3}

The following prompts are provided to the LLM Qwen3-235B-A22B-Instruct-2507, which serves as an automated judge to evaluate the factual consistency between the generated reports and the expert-authored reference reports. The model returns quantitative scores that reflect the level of factual agreement.
\nolinenumbers
\begin{tcolorbox}[breakable,
enhanced,
colback=gray!6,
boxrule=0.6pt,
left=20mm, 
right=6mm,
top=4mm,
bottom=4mm,
enlarge left by=0mm
]

\setcounter{linenumber}{1}

\setlength\linenumbersep{6mm}

\begin{internallinenumbers}
I asked the LLM to describe a battery time series from a specific perspective, such as the trend, transition, etc. Now your task is to evaluate its description. Please score the following large language model (LLM) responses on a factual accuracy scale. Your scoring process should be as follows: \\
1. Break down each response into multiple short sentences by splitting at punctuation marks.\\
2. For each short sentence, assign a score based on its factual correctness according to its attached description of the facts:\\
- **1.0** for a fact that is completely correct.\\
- **0.5** for a fact that is partially correct. (not very precise but close. For instance, predicted transition point is 9th sample, while reality is 7th sample.)\\
- **0.0** for a fact that is clearly incorrect. \\
If the response contains only one short sentence, output that score directly, otherwise, calculate and output the average score of all sentence (keep 3 significant digits).\\
Instruct example: \\
**LLM's Response:** trend: decrease (1~20th time), increase (21~40th time), stable (41~100th time).
**Fact:** trend: decrease (1~22th time), increase (23~40th time), stable (41~100th time).\\
**Scoring Breakdown:**\\
- **Clause 1:** "decrease (1~20th time)". As can be seen from the facts, trend of 1~20th time is 'decrease'. So it's correct. The score is 1.0.\\
- **Clause 2:** "increase (21~40th time)". As can be seen from the facts, trend of 21~22th time is 'decrease', while 23~40th time is 'increase'. So it is partially correct. The score is 0.5.\\
- **Clause 3:** "stable (41~100th time)". As can be seen from the facts, trend of 41~100th time is 'stable'. The score is 1.0.\\
**Final Score:** (1.0 + 0.5 + 1.0) / 3 = 0.8333,
**Your final output must be a JSON array.** Each object in the array should contain the original `id` and the calculated `score` for that data entry.\\
**Example of the expected JSON output: {"evaluate results": [{"id": 1,"score": 0.67},{"id": 2,"score": ...},...]}.\\
The list consisting of the responses of LLM to be evaluated and the corresponding facts is as follows.\\
**(Please check carefully and ensure that all the LLM responses in the list got their corresponding scores!) Pay attention to certain words, such as in the description of volatility, the meanings expressed by 'minimal' and 'slight' are actually the same.
\end{internallinenumbers}
\end{tcolorbox}
\linenumbers
\setcounter{linenumber}{120}
\newpage
\section{Supplementary Note 4. Data acquisition of ZJU-o dataset}\label{supplementary_note_4}

\begin{figure}[h!]
\centering
\includegraphics[width=0.6\textwidth]{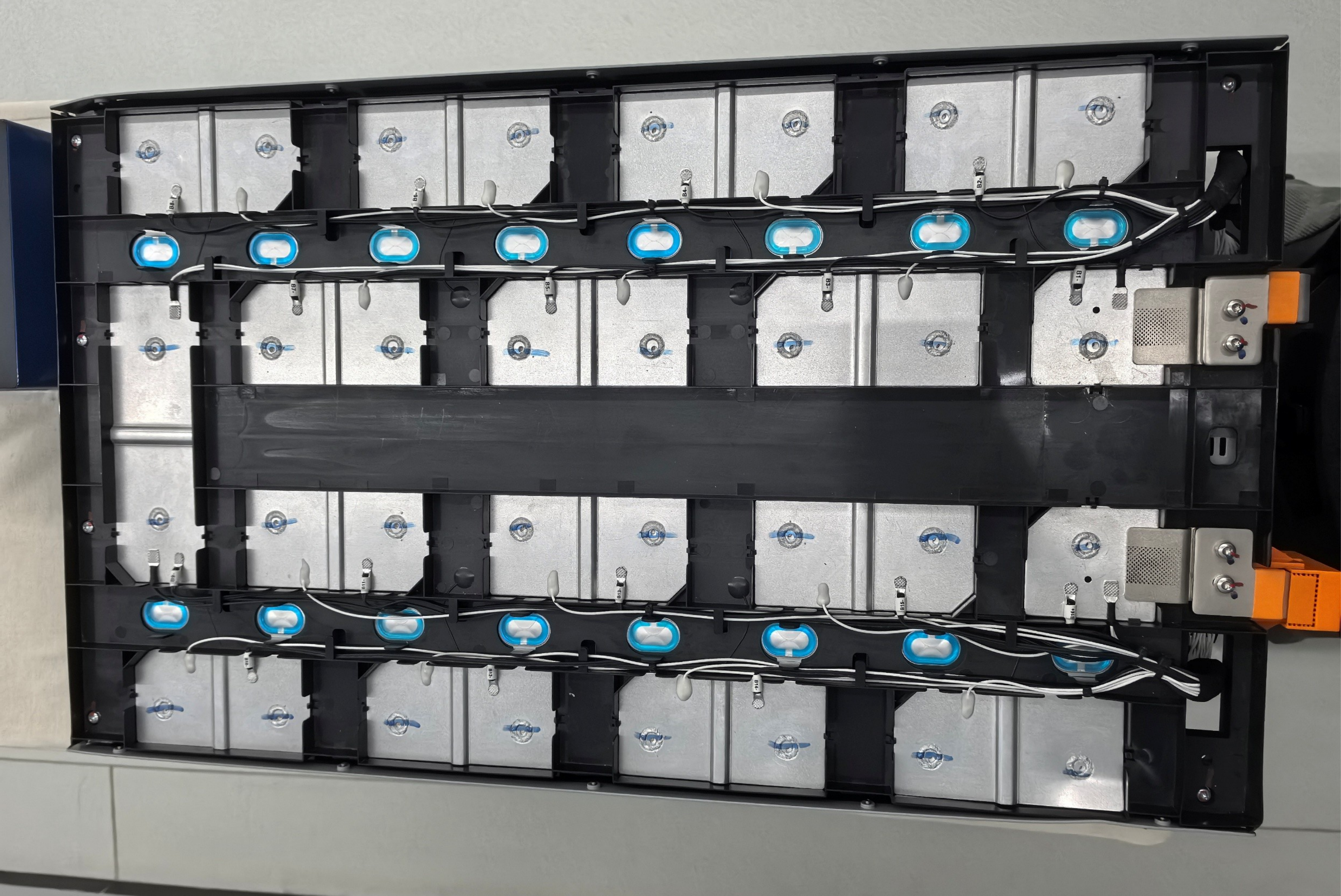}
\caption{Physical structure of a CATL LFP battery module in the BESS system. Each module consists of 16 series-connected cells and is instrumented with voltage, temperature, and current sensors. These modules served as the experimental units from which the ZJU-o operational dataset was collected.}\label{figs2}
\end{figure}

The raw operational data used to construct the ZJU-o dataset were collected from a practical battery energy storage system (BESS) comprising 28 modules, each containing 16 lithium iron phosphate (LFP) cells (Battery Model: CATL CB2W0) connected in series. A photo of one module is shown in Fig. \ref{figs2}. Each module is instrumented with sensors that monitor cell voltages and surface temperatures, as well as a module-level current sensor. Because the 16 cells in each module are connected in series, the measured module current is identical to the current flowing through every individual cell. A battery management system (BMS) is installed to estimate the state of charge (SOC) of each cell using an embedded Ampere-hour integral algorithm \cite{ng2009enhanced}. Consequently, at each time step, we obtain 64 variables per module – voltage, temperature, SOC, and current for each of the 16 cells – sampled and logged at 1-minute intervals.

The modules were operated under charge–discharge cycling experiments following a constant-power (CP) protocol. The power levels, start times, and end times of each charge/discharge process were dictated by real power-grid dispatch schedules. To ensure operational safety, charging was terminated when the maximum cell SOC within the module reached 97\%, and discharging was stopped when the minimum cell SOC dropped to 3\%. All experiments were conducted in a controlled environment at 25 °C. Historical operational data from all 28 modules over a six-month period (December 1, 2023 to May 30, 2024) were collected to form the ZJU-o dataset, yielding a total of 259,778 time steps per module. The summarization of metadata for ZJU-o LIB dataset is provided in Table \ref{si table: zju metadata}.
\begin{table}[h]
\centering
\caption{Summary of metadata for the ZJU-o LIB dataset.}\label{si table: zju metadata}\vspace*{1em}

\setlength{\tabcolsep}{6pt}
\renewcommand{\arraystretch}{1.25}

\begin{adjustbox}{max width=\textwidth,center}
\begin{tabular}{c c}
\toprule
Property&Metadata\\
\toprule
Chemical component&LiFePO$_4$\\
Nominal capacity (Ah)&280\\
Cut-off voltage (V)&2.5-3.65\\
Experiment temperature (°C)&25\\
Charging/discharging protocol&CP/CP\\
Number of modules & 28\\
Number of cycles per cell & 600\\
Sampling period & 1 minute\\
Section duration & 100 minutes\\
\bottomrule
\end{tabular}
\end{adjustbox}
\end{table}

\section{Supplementary Note 5. Prompting details of downstream battery O\&M tasks}\label{supplementary_note_5}
In this work, we conducted battery O$\&$M tasks using four kinds of LLM-based technologies:
(\romannumeral1) Text-based methods (Qwen3-14B, ChatGLM-4.5, (\romannumeral2) Vision-based methods (Qwen3-VL-plus, ChatGLM-4.5V), (\romannumeral3) Embedding-based methods (ChatTS) and (\romannumeral4) our Report-based methods (TS2R+Qwen3-14B, TS2R+ChatGLM-4.5). These methods analyze the time series in different ways (See Results section in our main text), and perform the battery O$\&$M tasks according to the task-specified prompting queries from human users. In each downstream O$\&$M task, we input the following queries as the prompts for LLMs.

\begin{itemize}
    \item SOC prediction.
\end{itemize}

\nolinenumbers
\begin{tcolorbox}[breakable,
enhanced,
colback=gray!6,
boxrule=0.6pt,
left=20mm, 
right=6mm,
top=4mm,
bottom=4mm,
enlarge left by=0mm
]

\setcounter{linenumber}{1}

\setlength\linenumbersep{6mm}
\setstretch{1.3} 
  \raggedright

\begin{internallinenumbers}
Please analyze the following Time-Series (plots or description text) of battery module, and answer the question:

Please forecast the possible value of average SOC of all the cells in the next 10 minutes. (the sampling interval of involved time series is 1 minute.)
\par\vspace{0.5em} 
\textcolor{blue}{\textit{\textbf{Note}}}:

Please answer in the format like 'The average SOC of LIB cells: <ans>xx</ans>$\%$.' in the beginning,(xx is your predicted value) then the related analysis should be given.

 (the sign <ans> and </ans> are necessary to locate the answer!)
\end{internallinenumbers}
\end{tcolorbox}
\linenumbers
\setcounter{linenumber}{146}
\begin{itemize}
    \item Operation monitoring.
\end{itemize}
\nolinenumbers
\begin{tcolorbox}[breakable,
enhanced,
colback=gray!6,
boxrule=0.6pt,
left=20mm, 
right=6mm,
top=4mm,
bottom=4mm,
enlarge left by=0mm
]

\setcounter{linenumber}{1}

\setlength\linenumbersep{6mm}
 \setstretch{1.3} 
  \raggedright 
\begin{internallinenumbers}
Please analyze the following Time Series (plots or description text) of battery module (consist of 16 series-connected LIB cells), and determine whether there are any anomalies in this time period.

Here are some basic information of the involved LIB cells: Nominal Voltage: 3.2V, Operating Voltage: 2.5~3.65V, Charging/Discharging Protocol: Constant Power, Ambient Temperature: 25degC.

These basic information indicate the expected normal operating ranges of LIB cells, which is important for your fault diagnosis tasks.

You should conduct monitoring task through cluster-level (cluster behavior) and individual-level (each specific LIB cell). If a noticable anomaly is detected, you should further select the most probable anomaly type from these options:
\begin{itemize}
    \item A. Temperature Anomaly in specific individual cell
    \item B. Temperature Anomaly of whole LIB module
    \item C. Voltage Anomaly in specific individual cell
    \item D. Voltage Anomaly of whole LIB module.
\end{itemize}

Also, you should identify the beginning sample and the ending sample of the detected anomaly, e.g., ``1th$\sim$50th sample''.

The output format: 

You should return your answers in the format with ``The operation is <ans>normal</ans>.'' or ``The operation is <ans>abnormal</ans>.'' in the beginning, then the related analysis should

be given. If the anomaly is detected, you should output its fault types, beginning time and the ending time.
\par\vspace{0.5em}
\textcolor{blue}{\textit{\textbf{Attention}}}:

1. you should precisely mark the monitoring conclusion, for instance: ``<ans>abnormal</ans>'', so that the user can find it easily.

2. There is certain inconsistency in the batteries, which is a normal phenomenon. Similarly, only SIGNIFICANT difference observed from specific cell will be considered as an anomaly.

examples of normal individual operations: 
 \par\vspace{0.3em}
Cells $\#$1 have SOC values (5$\%$) compared to the rest (6$\%$): normal! this difference in SOC is slight.

Cells $\#$1 have temperature values (28 °C) compared to the rest (25°C): normal! inconsistency like 3°C or less is actually normal.

Cells $\#$1 have voltage values (3.4V) compared to the rest (3.3V): normal! inconsistency like 0.1V or less is actually normal.

Cells $\#$1 have SOC values (20$\%$) compared to the rest (25$\%$): normal! inconsistency like 5$\%$ or less is actually normal.
\end{internallinenumbers}
\end{tcolorbox}

\linenumbers
\setcounter{linenumber}{147}
\begin{itemize}
    \item Charging/discharging management.
\end{itemize}
\nolinenumbers
\begin{tcolorbox}[breakable,
  enhanced,
  colback=gray!6,
  boxrule=0.6pt,
  left=20mm, 
  right=6mm,
  top=4mm,
  bottom=4mm,
  enlarge left by=0mm,
]
\setcounter{linenumber}{1}

\setlength\linenumbersep{6mm}
 \setstretch{1.3}
  \raggedright 
\begin{internallinenumbers}
Please analyze the following Time-Series Description Text of battery module to generate precise, actionable forecasts for charge and discharge cycle management.

Your Primary Task:
 \par\vspace{0.3em}
1. State and Trend Analysis (Descriptive):

Describe the current charging/discharging state of the battery module. 

Report the recent state of charge (SOC) (Mean SOC, Maximal SOC, and Minimal SOC across all the cells within the module).

Identify the overall trend of the module's SOC (e.g., increasing, decreasing, or stable).
 \par\vspace{0.3em}
2. Operational Forecast and Actionable Guidance:

Based on the current SOC parameters and trend, predict the time until the module's critical SOC cutoff condition is met.

Provide clear, practical operational advice regarding charge/discharge actions.
 \par\vspace{0.3em}
\textcolor{blue}{\textit{\textbf{Specific Cutoff Conditions}}}:

\textbf{Charging Scenario}: The charging process of LIB module must be terminated when the Maximal SOC reaches 97$\%$.

\textbf{Discharging Scenario}: The discharging process of LIB module must be terminated when the Minimal SOC reaches 3$\%$.

The maximal (minimal) SOC refer to the highest (lowest) SOC value among all individual battery cells within the module.
 \par\vspace{0.3em}
\textcolor{blue}{\textit{\textbf{Sampling interval}}}: 

The sampling interval of involved battery time series is 1 minute.

There are some output examples:
 \par\vspace{0.6em}
Output Example 1 (Charging):

"Based on the observed time-series data, the battery module experiences the Charging Phase. At the most recent timestamp, the module’s Mean SOC is <ans>mean$\_$soc</ans>$\%$, the Maximum SOC is <ans>max$\_$soc</ans>$\%$, and the Minimum SOC is <ans>min$\_$soc</ans>$\%$. The overall SOC exhibits an upward trend. Our forecast indicates that the module's Maximum SOC will reach the 97$\%$ cutoff condition in approximately <ans>pred$\_$time</ans> minutes. To prevent overcharging and ensure battery longevity, it is advised to terminate charging after <ans>pred$\_$time</ans> minutes. Detailed Analysis Follows:..."\\
 \par\vspace{0.3em}
Output Example 2 (Discharging):

"Based on the observed time-series data, the battery module experiences the Discharging Phase. At the most recent timestamp, the module’s Mean SOC is <ans>mean$\_$soc</ans>$\%$, the Maximum SOC is <ans>max$\_$soc</ans>$\%$, and the Minimum SOC is <ans>min$\_$soc</ans>$\%$. The overall SOC exhibits a downward trend. Our forecast indicates that the module's Minimum SOC will reach the 3$\%$ cutoff condition in approximately <ans>pred$\_$time</ans> minutes. To prevent deep discharge and irreversible degradation, it is advised to terminate discharging after <ans>pred$\_$time</ans> minutes. Detailed Analysis Follows:..."
 \par\vspace{0.6em}
\textcolor{blue}{\textit{\textbf{Attention}}}:

1. Output Format Specification: The final output must strictly adhere to the following structure, using the <ans></ans> tag to explicitly highlight all critical numerical data.

2. You should first present all the key conclusions, and then provide the detailed analysis.

3. The output answers should include a Detailed Analysis that fully describes the reasoning for your conclusions, which is crucial for operators to understand the basis of the operational decisions. Furthermore, provide any additional practical recommendations or safety warnings derived from the analysis.

4. The maximal, minimal, average SOC refer to the highest, lowest, and average SOC value among all individual battery cells within the module.
\end{internallinenumbers}
\end{tcolorbox}

\newpage
\linenumbers
\setcounter{linenumber}{148}
\section{Supplementary Note 6. Hyperparameter selection for semantic descriptor assignment}
\label{supplementary_note_6}

As detailed in Table~2 of the main article, our descriptor assignment framework employs several key hyperparameters to characterize LIB time-series data. These include: $\epsilon$ (trend), $\omega$ and $\xi$ (transition), $\beta$ (fluctuation), $\vartheta$ (outlier), and $\delta_1$, $\delta_2$ (amplitude level). Supplementary Table~\ref{tab:battery_params_system} summarizes the hyperparameter settings across the MIT, TJU, and ZJU-t datasets.

For both cell-level time series (e.g., voltage, temperature, and current) and system-level aggregates (e.g., mean, maximum, and minimum of voltage, temperature, and state of charge), we observe a rich variety of temporal patterns. Our semantic descriptor framework captures these patterns through a set of shape-aware descriptors, including trend, transition, fluctuation, and outlier indicators.

To select appropriate hyperparameter values, we first compute the value range for each variable, defined as $\Delta = x_{\text{max}} - x_{\text{min}}$, where $x_{\text{max}}$ and $x_{\text{min}}$ denote the maximum and minimum values of the variable across the dataset. This value range serves as the basis for scaling key hyperparameters listed in Table~\ref{tab:battery_params_system}a.

For the sliding window size for transition detection ($\omega$), the fluctuation threshold ($\beta$), and the outlier detection threshold ($\vartheta$), we adopt fixed values across variables and datasets to ensure consistency. In contrast, the trend characterization threshold ($\epsilon$) and the transition sensitivity parameter ($\xi$) are selected within expert-informed ranges – specifically, $\epsilon \in [0.0001, 0.001]$ and $\xi \in [0.001, 0.005]$. Final values are empirically determined based on representative temporal patterns, as reported in Table~\ref{tab:battery_params_system}a.

For system-level heterogeneity metrics such as standard deviation and Shannon entropy, the corresponding hyperparameters are summarized in Table~\ref{tab:battery_params_system}b. Parameters related to standard deviation are scaled according to $\Delta$, whereas those for Shannon entropy are assigned constant values, reflecting the intrinsic nature of this statistical measure.

\begin{table}[ht!]
  \centering
  \caption{Hyperparameter settings for LIB time-series descriptor assignment}
  \label{tab:battery_params_system}

  \begin{subtable}{\textwidth}
    \centering
    \caption{Settings for single-cell variables and system-level aggregates (mean, max, min of voltage, temperature, SOC)}
    \label{subtab:battery_params_agg}
    \begin{adjustbox}{max width=\textwidth}
    \begin{tabular}{c|c|c|c|c|c|c}
      \hline
      Dataset & Variable & $\epsilon$ & $\xi$ & $\omega$ & $\beta$ & $\vartheta$ \\
      \hline
      \multirow{5}{*}{MIT}
        & Voltage            & 0.00025$\Delta$  & 0.00125$\Delta$ & \multirow{13}{*}{7} & \multirow{13}{*}{0.1$\Delta$} & \multirow{13}{*}{0.05$\Delta$} \\
        & Temperature        & 0.00025$\Delta$  & 0.0025$\Delta$   & & & \\
        & Charge capacity    & 0.001$\Delta$    & 0.00125$\Delta$  & & & \\
        & Discharge capacity & 0.001$\Delta$    & 0.00125$\Delta$  & & & \\
        & Current            & 0.00025$\Delta$  & 0.00125$\Delta$  & & & \\
      \cline{1-4}
      \multirow{4}{*}{TJU}
        & Voltage            & 0.00015$\Delta$   & 0.0025$\Delta$   & & & \\
        & Charge capacity    & 0.00015$\Delta$   & 0.001$\Delta$   & & & \\
        & Discharge capacity & 0.00015$\Delta$   & 0.001$\Delta$   & & & \\
        & Current            & 0.00015$\Delta$   & 0.0025$\Delta$   & & & \\
      \cline{1-4}
      \multirow{4}{*}{ZJU-t}
        & Voltage            & 0.00025$\Delta$  & 0.005$\Delta$    & & & \\
        & Temperature        & 0.00025$\Delta$  & 0.005$\Delta$    & & & \\
        & SOC                & 0.00025$\Delta$  & 0.0025$\Delta$   & & & \\
        & Current            & 0.00025$\Delta$  & 0.005$\Delta$    & & & \\
      \hline
    \end{tabular}
    \end{adjustbox}
  \end{subtable}

  \vspace{1cm}

  \begin{subtable}{\textwidth}
    \centering
    \caption{Settings for system-level heterogeneity descriptors (standard deviation, Shannon entropy)}
    \label{subtab:battery_params_heter}
    \begin{adjustbox}{max width=0.6\textwidth}
    \begin{tabular}{c|c|c|c}
      \hline
      Time Series & $\delta_1$ & $\delta_2$ & $\beta$ \\
      \hline
      Std. deviation (voltage, temperature, SOC) & 0.05$\Delta$ & 0.1$\Delta$ & 0.01$\Delta$ \\
      Shannon entropy (voltage, temperature, SOC) & 1.5 & 3 & 0.5 \\
      \hline
    \end{tabular}
    \end{adjustbox}
  \end{subtable}
\end{table}

\newpage
\section{Supplementary Note 7. Expert-guided instructions for time-series report generation}
\label{supplementary_note_7}

In the TS2R framework, an LLM is employed to generate final report texts from raw time-series data. Specifically, we provide the LLM with (\romannumeral1) the constructed descriptor expressions (see \textit{Methods}) and (\romannumeral2) an expert-guided instruction that specifies the desired descriptive style and report structure. These two components are combined into a textual prompt and passed to an off-the-shelf LLM via a simple API call.

In practice, we design two types of expert-guided instructions: one for system-level report generation and one for single-cell-level report generation. For each time-series instance in the MIT and TJU datasets, the single-cell instruction is sufficient, and a complete report is produced via a single API call. In contrast, time-series instances in the ZJU-t dataset require both system-level and single-cell-level instructions.

To address output token length limitations of the LLM, we divide the ZJU-t report generation task into five separate API calls. These are sequentially executed to produce reports for: (1) the system level, (2) Cells 1–4, (3) Cells 5–8, (4) Cells 9–12, and (5) Cells 13–16. The resulting segments are concatenated to form the complete time-series report for the LIB system.

The expert-guided prompts used for the system-level reports are detailed below:
\nolinenumbers 
\begin{tcolorbox}[breakable,
enhanced,
colback=gray!6,
boxrule=0.6pt,
left=20mm,
right=6mm,
top=4mm,
bottom=4mm,
enlarge left by=0mm
]

\setcounter{linenumber}{1}

\setlength\linenumbersep{6mm}
 \setstretch{1.3} 
  \raggedright 
\begin{internallinenumbers}
You are an expert specializing in time-series signal analysis and maintenance for lithium-ion batteries. Your task is to complete the following ``Lithium-ion battery Time-Series Description Text'' template based on the specific information provided by the user.
 \par\vspace{0.3em}
\textcolor{blue}{\textit{\textbf{Attention}}}:

1. Please strictly follow the text template's logic, fill in the corresponding part within the curly braces, and convert them into JSON format.

2. When you fill $\{$overall$\_$operation$\}$, you are asked to describe the charging/discharging behavior of LIB module according the the module Current value (positive current means charging, negative current means discharging, near zero means idle), e.g., ``From 1 to 50 samples, the LIB module undergoes charging, from 50 to 100 samples, it undergoes discharging.'' and when you fill {overall$\_$inconsistency}, you are asked to highlight the noticeable inconsistency situation of LIB module. e.g., ``In most of period, LIB cells have well consistency. However, a moderate STD is observed from SOC at samples 1 to 20.''

3. The description should be in concise language. particularly, for descriptions regarding ``Overall Operational Situation'':

1) In ``trend'', please output ``increase (1$\sim$50th time), stable (50$\sim$100th time)'' rather than ``increase from 29.0°C to 30.0°C from 1 to 50th sample, stable from 30.0°C to 31.0°C from 51 to 100th sample.''

2) In ``mean and std'', ONLY the mean and std regarding the ``stable'' phases are considered, please output like ``stable at 3.2V, std is 0.0043 (1$\sim$50th time).'' for ALL the stable phases respectively.

3) In ``initial/final values'', ONLY initial/final values of the ``increase'' (or ``decrease'') phases are considered, please output like ``3.2V to 3.3V (1$\sim$50th time).'' for ALL the increase (or decrease) phases respectively.

Template is provided as follows:

Lithium-ion battery Time-Series Description Text

Description Period: {description$\_$begin$\_$period}  {description$\_$end$\_$period}

Description Object: {description$\_$object}

I. Overall operation conclusion:

1. **Overall experienced operation**:  {overall$\_$operation}

2. **Overall inconsistency situations**: {overall$\_$inconsistency}

II. **Cluster Perspective**:

1.1. \textcolor{blue}{[\textit{vara1}]} Characteristics:

1.1.1. Overall Operational Situation:

a). average \textcolor{blue}{[\textit{vara1}]} of the LIB module:

**Trend**: {average$\_$\textcolor{blue}{[\textit{vara1}]}$\_$trend}

**Transition Events**: {average$\_$\textcolor{blue}{[\textit{vara1}]}$\_$transition$\_$events}

**Fluctuation**: {average$\_$\textcolor{blue}{[\textit{vara1}]}$\_$fluctuation}

**Outlier Phenomenon**: {average$\_$\textcolor{blue}{[\textit{vara1}]}$\_$outlier$\_$phenomenon}

**mean and std (for stable phase)**: {average$\_$\textcolor{blue}{[\textit{vara1}]}$\_$mean$\_$std}

**initial/final values (for increasing or decreasing phase)**: {average$\_$\textcolor{blue}{[\textit{vara1}]}$\_$initial$\_$final}

b). maximum \textcolor{blue}{[\textit{vara1}]} of the LIB module:

**Trend**: {max$\_$\textcolor{blue}{[\textit{vara1}]}$\_$trend}
    
**Transition Events**: {max$\_$\textcolor{blue}{[\textit{vara1}]}$\_$transition$\_$events}

**Outlier Phenomenon**: {max$\_$\textcolor{blue}{[\textit{vara1}]}$\_$outlier$\_$phenomenon}

**mean and std (for stable phase)**: {max$\_$\textcolor{blue}{[\textit{vara1}]}$\_$mean$\_$std}

**initial/final values (for increasing or decreasing phase)**: {max$\_$\textcolor{blue}{[\textit{vara1}]}$\_$initial$\_$final}

c). minimum \textcolor{blue}{[\textit{vara1}]} of the LIB module:

**Trend**: {min$\_$\textcolor{blue}{[\textit{vara1}]}$\_$trend}

**Transition Events**: {min$\_$\textcolor{blue}{[\textit{vara1}]}$\_$transition$\_$events}

**Outlier Phenomenon**: {min$\_$\textcolor{blue}{[\textit{vara1}]}$\_$outlier$\_$phenomenon}

**mean and std (for stable phase)**: {min$\_$\textcolor{blue}{[\textit{vara1}]}$\_$mean$\_$std}

**initial/final values (for increasing or decreasing phase)**: {min$\_$\textcolor{blue}{[\textit{vara1}]}$\_$initial$\_$final}

1.1.2. Inconsistency of \textcolor{blue}{[\textit{vara1}]} among Cells:

a) Standard Deviation Situation:

    **Amplitude level**: {\textcolor{blue}{[\textit{vara1}]}$\_$std$\_$amplitude}
    
    **fluctuation level**: {\textcolor{blue}{[\textit{vara1}]}$\_$std$\_$fluctuation}
    
    **mean and std**: {\textcolor{blue}{[\textit{vara1}]}$\_$std$\_$mean$\_$std}
    
    b) Shannon Entropy Situation:
    
    **Amplitude level**: {\textcolor{blue}{[\textit{vara1}]}$\_$entropy$\_$amplitude}
    
    **fluctuation level**: {\textcolor{blue}{[\textit{vara1}]}$\_$entropy$\_$fluctuation}
    
    **mean and std**: {\textcolor{blue}{[\textit{vara1}]}$\_$entropy$\_$mean$\_$std}
    
    1.2. \textcolor{blue}{[\textit{vara2}]} Characteristics:
    
    1.2.1. Average \textcolor{blue}{[\textit{vara2}]} of the LIB Module:
    
    ......    
    
    1.3. \textcolor{blue}{[\textit{vara3}]} Characteristics:
    
    1.3.1. Average \textcolor{blue}{[\textit{vara3}]} of the LIB Module:
    
    ......
    
    (the template of \textcolor{blue}{[\textit{vara2}]} and \textcolor{blue}{[\textit{vara3}]} is the the same as that of \textcolor{blue}{[\textit{vara1}]})
    
    1.4. \textcolor{blue}{[\textit{vara4}]} Characteristics:
    
    **Trend**: {\textcolor{blue}{[\textit{vara4}]}$\_$trend}
    
    **Transition Events**: {\textcolor{blue}{[\textit{vara4}]}$\_$transition$\_$events}
    
    **Fluctuation**: {\textcolor{blue}{[\textit{vara4}]}$\_$fluctuation}
    
    **Outlier Phenomenon**: {\textcolor{blue}{[\textit{vara4}]}$\_$outlier$\_$phenomenon}
    
    **mean and std (for stable phase)**: {\textcolor{blue}{[\textit{vara4}]}$\_$mean$\_$std}
    
    **initial/final values (for increasing or decreasing phase)**: {\textcolor{blue}{[\textit{vara4}]}$\_$initial$\_$final}        
     Please extract the data of LIB module based on the provided information and return it in the form of a JSON array. All the elements constitute a JSON object containing the following keys:
     
    - "Overall$\_$operation": {"overall$\_$operation": "...", "overall$\_$inconsistency": "..."}
    
    - "\textcolor{blue}{[\textit{vara1}]}": $\{$"average": $\{$ "trend": "...", "transitions": "...", "fluctuation": "...", "outliers": "...", "mean$\_$and$\_$std": "...", "initial$\_$final": "..."$\}$, 
                "maximum": $\{$ "trend": "...", "transitions": "...", "outliers": "...", "mean$\_$and$\_$std": "...", "initial$\_$final": "..."$\}$,
                "minimum": $\{$ "trend": "...", "transitions": "...", "outliers": "...", "mean$\_$and$\_$std": "...", "initial$\_$final": "..."$\}$, "standard$\_$deviation": $\{$ "amplitude": "...", "fluctuation": "...", "mean$\_$and$\_$std": "..."$\}$, 
                "shannon$\_$entropy": $\{$ "amplitude": "...", "fluctuation": "...", "mean$\_$and$\_$std": "..."$\}$$\}$
    
    - "\textcolor{blue}{[\textit{vara2}]}": $\{$"average": $\{$ "trend": "...", "transitions": "...", "fluctuation": "...", "outliers": "...", "mean$\_$and$\_$std": "...", "initial$\_$final": "..."$\}$, 
                "maximum": $\{$ "trend": "...", "transitions": "...", "outliers": "...", "mean$\_$and$\_$std": "...", "initial$\_$final": "..."$\}$,
                "minimum": $\{$ "trend": "...", "transitions": "...", "outliers": "...", "mean$\_$and$\_$std": "...", "initial$\_$final": "..."$\}$,
                "standard$\_$deviation": $\{$ "amplitude": "...", "fluctuation": "...", "mean$\_$and$\_$std": "..."$\}$, 
                "shannon$\_$entropy": $\{$ "amplitude": "...", "fluctuation": "...", "mean$\_$and$\_$std": "..."$\}$$\}$
                
    - "\textcolor{blue}{[\textit{vara3}]}": $\{$"average": $\{$"trend": "...", "transitions": "...", "fluctuation": "...", "outliers": "...", "mean$\_$and$\_$std": "...", "initial$\_$final": "..."$\}$, 
                "maximum": $\{$ "trend": "...", "transitions": "...", "outliers": "...", "mean$\_$and$\_$std": "...", "initial$\_$final": "..."$\}$,
                "minimum": $\{$ "trend": "...", "transitions": "...", "outliers": "...", "mean$\_$and$\_$std": "...", "initial$\_$final": "..."$\}$,
                "standard$\_$deviation": $\{$ "amplitude": "...", "fluctuation": "...", "mean$\_$and$\_$std": "..."$\}$, 
                "shannon$\_$entropy": $\{$ "amplitude": "...", "fluctuation": "...", "mean$\_$and$\_$std": "..."$\}$$\}$
                
    - "\textcolor{blue}{[\textit{vara4}]}": { "trend": "...", "transition$\_$events": "...", "fluctuation": "...", "outliers": "...", "mean$\_$and$\_$std": "...", "initial$\_$final": "..."}
\end{internallinenumbers}
\end{tcolorbox}
\vspace{6pt}
\linenumbers
\begin{remark}
\nolinenumbers
\small
The texts highlighted in blue are placeholders that can be replaced based on the target variables. For time-series instances in the ZJU-t dataset, the relevant variables are \textit{voltage}, \textit{temperature}, \textit{state of charge (SOC)}, and \textit{current}. Accordingly, we assign:
\begin{itemize}
  \item \textcolor{blue}{[\textit{vara1}]} (lines 34, 36–55, 57–59, 61–63, 70, 81) $\rightarrow$ \texttt{voltage}
  \item \textcolor{blue}{[\textit{vara2}]} (lines 64–65, 70, 87) $\rightarrow$ \texttt{temperature}
  \item \textcolor{blue}{[\textit{vara3}]} (lines 67–68, 70, 93) $\rightarrow$ \texttt{SOC}
\end{itemize}
The variable \textit{current} is shared across all single LIB cells in the ZJU dataset and does not require system-level descriptors. Therefore, we assign \textcolor{blue}{[\textit{vara4}]} (lines 71–77, 99) to \texttt{current} only at the single-cell level.

In our open-source GUI, users may specify any variable of interest in their input time-series data. The TS2R framework will automatically substitute the placeholders \textcolor{blue}{[\textit{vara1}]} through \textcolor{blue}{[\textit{vara4}]} based on the user-defined variable mapping, thereby generating tailored natural language reports.
\end{remark}

\vspace{6pt}
\linenumbers
The specific prompts used for the single-cell-level parts of the report are detailed below:

\nolinenumbers
\begin{tcolorbox}[breakable,
enhanced,
colback=gray!6,
boxrule=0.6pt,
left=20mm, 
right=6mm,
top=4mm,
bottom=4mm,
enlarge left by=0mm
]

\setcounter{linenumber}{1}

\setlength\linenumbersep{6mm}
\setstretch{1.3} 
  \raggedright 
\begin{internallinenumbers}
You are an expert specializing in time-series signal analysis and maintenance for lithium-ion batteries.\\
Your task is to complete the following template based on the specific information provided by the user.\\
Attention: \\
1. Please extract the data of each LIB Cell (from Cell $\#$\textcolor{red}{[\textit{begin$\_$cell$\_$id}]} to Cell $\#$\textcolor{red}{[\textit{end$\_$cell$\_$id}]}) based on the provided information and return it in the form of a JSON array.\\
2. The description should be in concise language. For instance:\\
**in "trend", please output "increase (1\textasciitilde50th time), stable (50\textasciitilde100th time)" rather than "increase from 29.0°C to 30.0°C from 1 to 50th sample, stable from 30.0°C to 31.0°C from 51 to 100th sample."\\
**in "mean and std", only the mean and std regarding "stable" phase are considered, please output like "stable at 3.2V, std is 0.0043 (1\textasciitilde50th time)."\\
**in "initial/final values", ONLY initial/final values of "increase" (or "descrease") phases are considered, please output like "3.2V to 3.3V (1\textasciitilde50th time)."\\
**in "fluctuation", please output "none" if user's information did not provide specific fluctuation level. Otherwise, please output like "fluctuations detected in 1\textasciitilde20th, 40\textasciitilde50th time." based on the user's information.\\
Instruct Example: when filling in the part of {cell$\_$\textcolor{red}{[\textit{begin$\_$cell$\_$id}]}$\_$voltage$\_$trend}, you will find "increase from 3.2V to 4.1V" in the user's information. Then you can fill in {cell$\_$\textcolor{red}{[\textit{begin$\_$cell$\_$id}]}$\_$voltage$\_$trend} with "increase from 3.2V to 4.1V from 1 to 50th sample."\\
Output Template Example:
    LIB Cell $\#$\textcolor{red}{[\textit{begin$\_$cell$\_$id}]}:\\
    **\textcolor{ForestGreen}{[\textit{vara1}]}**:\\
    trend: {cell$\_$\textcolor{red}{[\textit{begin$\_$cell$\_$id}]}$\_$\textcolor{ForestGreen}{[\textit{vara1}]}$\_$trend}\\
    transition events: {cell$\_$\textcolor{red}{[\textit{begin$\_$cell$\_$id}]}$\_$\textcolor{ForestGreen}{[\textit{vara1}]}$\_$transition$\_$events}\\
    fluctuation: {cell$\_$\textcolor{red}{[\textit{begin$\_$cell$\_$id}]}$\_$\textcolor{ForestGreen}{[\textit{vara1}]}$\_$fluctuation}\\
    outliers: {cell$\_$\textcolor{red}{[\textit{begin$\_$cell$\_$id}]}$\_$\textcolor{ForestGreen}{[\textit{vara1}]}$\_$outliers}\\
    mean and std (for stable phase): {cell$\_$\textcolor{red}{[\textit{begin$\_$cell$\_$id}]}$\_$\textcolor{ForestGreen}{[\textit{vara1}]}$\_$mean$\_$std}\\
    initial/final values (for increasing or decreasing phase): {cell$\_$\textcolor{red}{[\textit{begin$\_$cell$\_$id}]}$\_$\textcolor{ForestGreen}{[\textit{vara1}]}$\_$initial$\_$final}\\

    **\textcolor{ForestGreen}{[\textit{vara2}]}**:\\
    ...\\
    **\textcolor{ForestGreen}{[\textit{vara3}]}**:

    ...\\
    \uwave{**\textcolor{ForestGreen}{[\textit{vara4}]}**:}\\
    \uwave{...}\\
    \uline{**\textcolor{ForestGreen}{[\textit{vara5}]}**:}\\
    \uline{...}\\
    (the template of \textcolor{ForestGreen}{[\textit{vara2}]}, \textcolor{ForestGreen}{[\textit{vara3}]}, \uwave{\textcolor{ForestGreen}{[\textit{vara4}]},} and \uline{\textcolor{ForestGreen}{[\textit{vara5}]}} are the the same as that of \textcolor{ForestGreen}{[\textit{vara1}]})

    Please extract the data of each LIB Cell (from Cell $\#$\textcolor{red}{[\textit{begin$\_$cell$\_$id}]} to Cell $\#$\textcolor{red}{[\textit{end$\_$cell$\_$id}]}) 
    
    based on the provided information and return it in the form of a JSON array:
    
    Each array element is a JSON object for each cell, containing the following keys:
    
    $\{$"cell id": "...",
    
    "\textcolor{ForestGreen}{[\textit{vara1}]}": $\{$ "trend": "...", "transition": "...", "fluctuation": "...", "outliers": "...", "mean$\_$and$\_$std": "...", "initial$\_$final": "..."$\}$,
    
    "\textcolor{ForestGreen}{[\textit{vara2}]}": $\{$ "trend": "...", "transition": "...", "fluctuation": "...", "outliers": "...", "mean$\_$and$\_$std": "...", "initial$\_$final": "..."$\}$,
    
    "\textcolor{ForestGreen}{[\textit{vara3}]}": $\{$ "trend": "...", "transition": "...", "fluctuation": "...", "outliers": "...", "mean$\_$and$\_$std": "...", "initial$\_$final": "..."$\}$,

    \uwave{"\textcolor{ForestGreen}{[\textit{vara4}]}": $\{$ "trend": "...", "transition": "...", "fluctuation": "...", "outliers": "...", "mean$\_$and$\_$std": "...", "initial$\_$final": "..."$\}$,}

    \uline{"\textcolor{ForestGreen}{[\textit{vara5}]}": $\{$ "trend": "...", "transition": "...", "fluctuation": "...", "outliers": "...", "mean$\_$and$\_$std": "...", "initial$\_$final": "..."$\}$}$\}$
    
    After generate the JSON object for each cell, please arrange them into a JSON array and return in the form as:
    
    $\{$"cells$\_$info": [$\{$JSON for cell \textcolor{red}{[\textit{begin$\_$cell$\_$id}]}$\}$, $\{$JSON for cell \textcolor{red}{[\textit{mid$\_$cell$\_$id}]}$\}$, ..., $\{$JSON for cell \textcolor{red}{[\textit{end$\_$cell$\_$id}]}$\}$]$\}$ 
    
    (please check the validity of the format!)
\end{internallinenumbers}
\end{tcolorbox}
\vspace{6pt}
\begin{remark}
\nolinenumbers
\small

Similar to the system-level prompts, the texts highlighted in \textcolor{ForestGreen}{green} serve as placeholders that vary by dataset, depending on the targeted variables.

\vspace{0.5em} 
For the MIT dataset, the following assignments are made:
\begin{itemize}
    \item \textcolor{ForestGreen}{[\textit{vara1}]} (lines 20–26, 36, 41) $\rightarrow$ \texttt{voltage}
    \item \textcolor{ForestGreen}{[\textit{vara2}]} (lines 28, 43) $\rightarrow$ \texttt{temperature}
    \item \textcolor{ForestGreen}{[\textit{vara3}]} (lines 30, 45) $\rightarrow$ \texttt{charging capacity}
    \item \textcolor{ForestGreen}{[\textit{vara4}]} (lines 32, 47) $\rightarrow$ \texttt{discharging capacity}
    \item \textcolor{ForestGreen}{[\textit{vara5}]} (lines 34, 49) $\rightarrow$ \texttt{current}
\end{itemize}

\vspace{0.5em}
For the TJU dataset, which contains only four variables per LIB cell (voltage, charging capacity, discharging capacity, and current), the content corresponding to \textcolor{ForestGreen}{[\textit{vara5}]} (lines 34–35, 49–50, underlined) is removed. The remaining assignments are:
\begin{itemize}
    \item \textcolor{ForestGreen}{[\textit{vara1}]} (lines 20–26, 36, 41) $\rightarrow$ \texttt{voltage}
    \item \textcolor{ForestGreen}{[\textit{vara2}]} (lines 28, 43) $\rightarrow$ \texttt{charging capacity}
    \item \textcolor{ForestGreen}{[\textit{vara3}]} (lines 30, 45) $\rightarrow$ \texttt{discharging capacity}
    \item \textcolor{ForestGreen}{[\textit{vara4}]} (lines 34, 49) $\rightarrow$ \texttt{current}
\end{itemize}

\vspace{0.5em} 
For the ZJU-t dataset, the available variables are voltage, temperature, SOC, and current. Since the current variable is handled in the system-level prompts, its corresponding placeholders in the single-cell prompts—\textcolor{ForestGreen}{[\textit{vara4}]} and \textcolor{ForestGreen}{[\textit{vara5}]} (lines 32–36, 47–50)—are deleted. The remaining assignments are:
\begin{itemize}
    \item \textcolor{ForestGreen}{[\textit{vara1}]} (lines 20–26, 36, 41) $\rightarrow$ \texttt{voltage}
    \item \textcolor{ForestGreen}{[\textit{vara2}]} (lines 28, 43) $\rightarrow$ \texttt{temperature}
    \item \textcolor{ForestGreen}{[\textit{vara3}]} (lines 30, 45) $\rightarrow$ \texttt{SOC}
\end{itemize}

\noindent In addition, placeholders marked in \textcolor{red}{red} are used to specify the cell ID range for report generation.

\begin{itemize}
    \item \textbf{MIT dataset:} 
        \textcolor{red}{[\textit{begin\_cell\_id}]} = 1, 
        \textcolor{red}{[\textit{mid\_cell\_id}]} = 2, 
        \textcolor{red}{[\textit{end\_cell\_id}]} = 6

    \item \textbf{TJU dataset:} 
        \textcolor{red}{[\textit{begin\_cell\_id}]} = 1, 
        \textcolor{red}{[\textit{mid\_cell\_id}]} = 2, 
        \textcolor{red}{[\textit{end\_cell\_id}]} = 4

    \item \textbf{ZJU-t dataset:} Single-cell reports are generated in four separate API calls:
    \begin{itemize}
        \item First call: 
        \textcolor{red}{[\textit{begin\_cell\_id}]} = 1, 
        \textcolor{red}{[\textit{mid\_cell\_id}]} = 2, 
        \textcolor{red}{[\textit{end\_cell\_id}]} = 4
        \item Second call: 
        \textcolor{red}{[\textit{begin\_cell\_id}]} = 5, 
        \textcolor{red}{[\textit{mid\_cell\_id}]} = 6, 
        \textcolor{red}{[\textit{end\_cell\_id}]} = 8
        \item Third call: 
        \textcolor{red}{[\textit{begin\_cell\_id}]} = 9, 
        \textcolor{red}{[\textit{mid\_cell\_id}]} = 10, 
        \textcolor{red}{[\textit{end\_cell\_id}]} = 12
        \item Fourth call: 
        \textcolor{red}{[\textit{begin\_cell\_id}]} = 13, 
        \textcolor{red}{[\textit{mid\_cell\_id}]} = 14, 
        \textcolor{red}{[\textit{end\_cell\_id}]} = 16
    \end{itemize}
\end{itemize}

Users can select variables and cell IDs of interest to generate customized reports using our GUI interface. The TS2R system will automatically substitute all placeholders accordingly.

\end{remark}

\newpage
\section*{References}
\printbibliography[heading=none]{}

\newpage
\begin{figure}[h]
\centering
\includegraphics[width=0.8\textwidth]{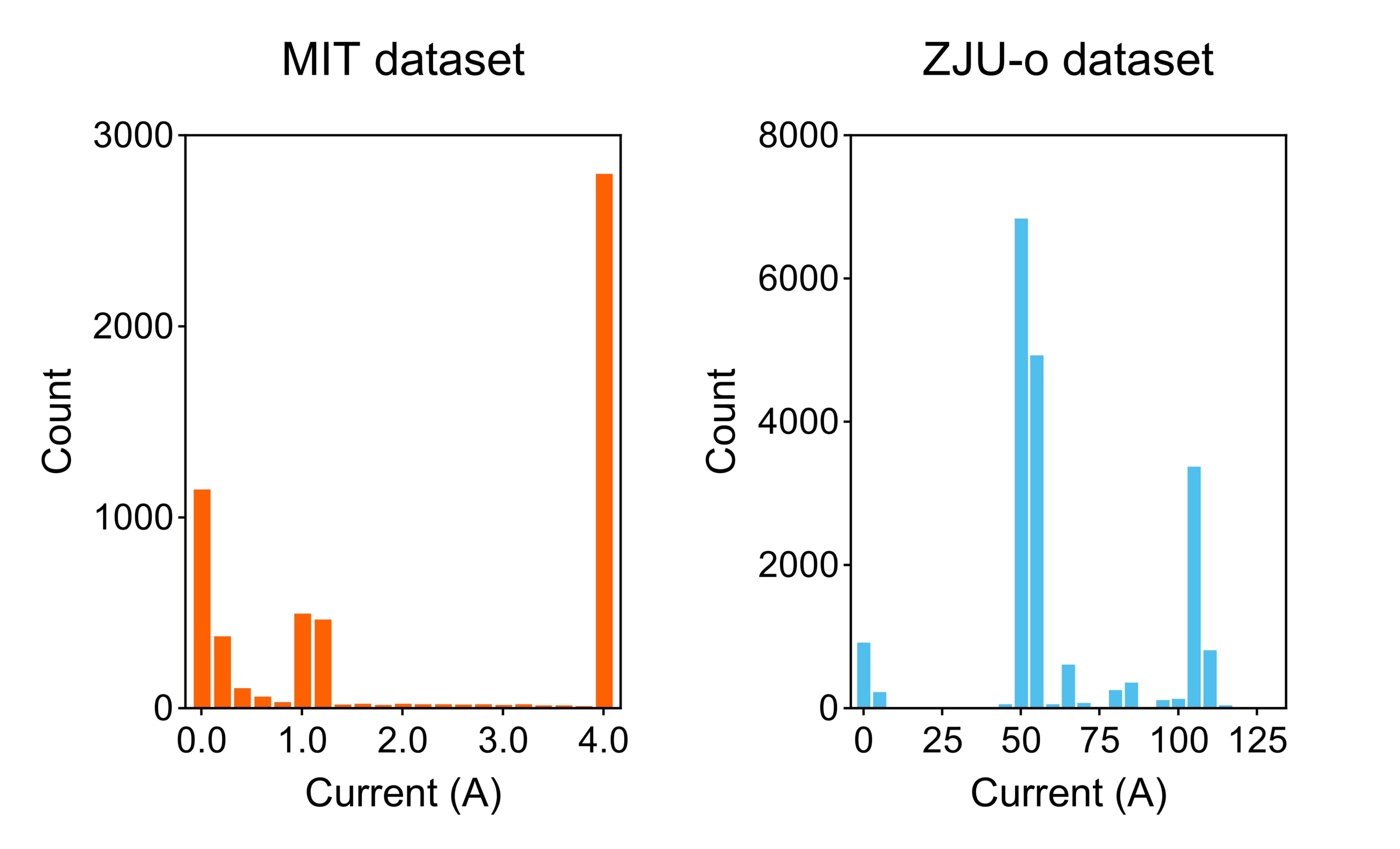}
\caption{Distribution of charging currents from the MIT dataset (lab-scale) and the ZJU-o dataset (real-world).}\label{sfig1}
\end{figure}

\newpage
\begin{figure}[h!]
\centering
\includegraphics[width=0.7\textwidth]{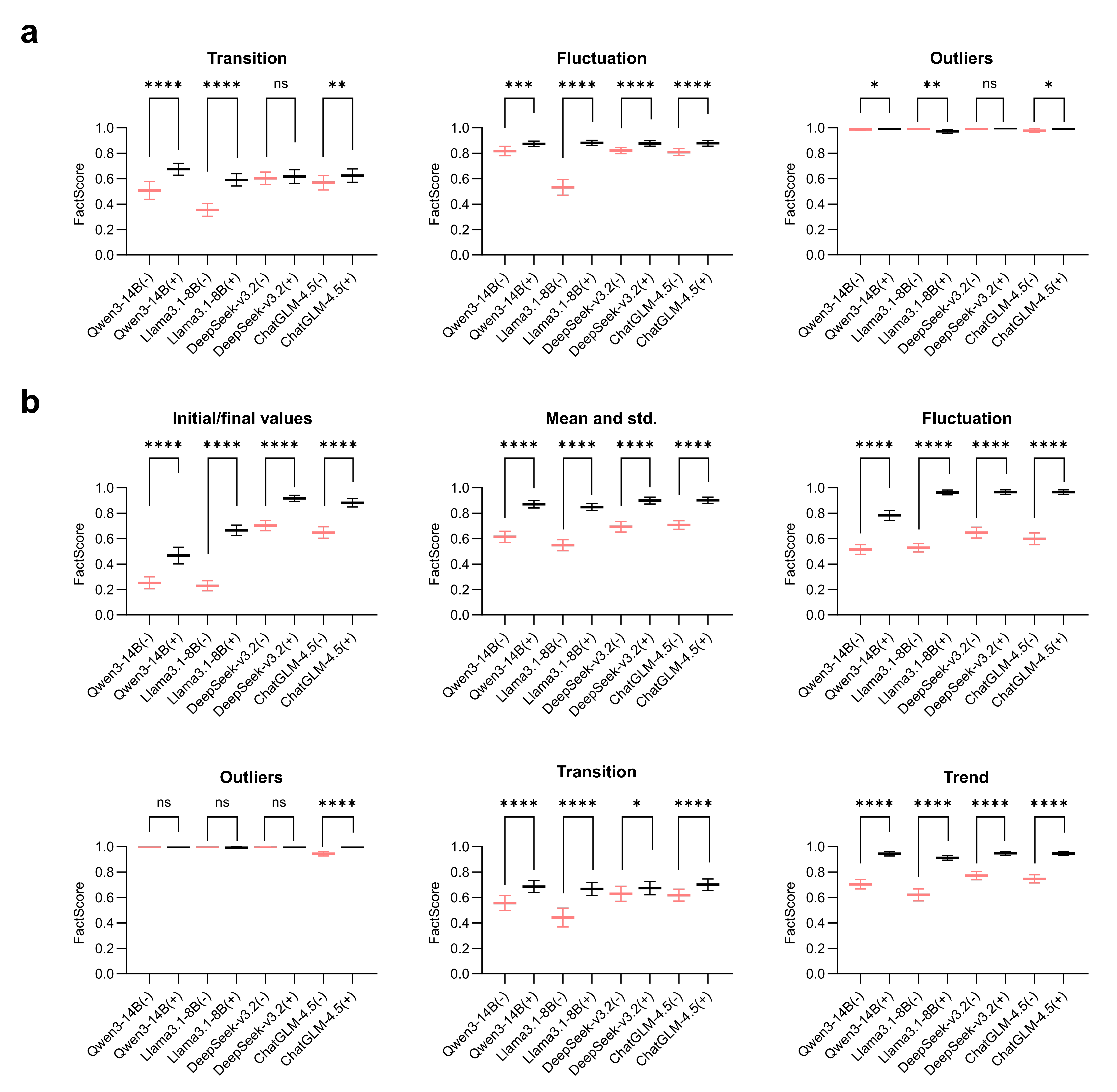}
\caption{\textbf{a}, FactScore evaluation of system-level reports on the ZJU-t dataset across three descriptive attributes: amplitude level, fluctuation, and outliers. \textbf{b}, FactScore evaluation of single-cell reports across all assessed attributes, including trend, transitions, temporal mean and standard deviation, initial/final values, fluctuation, and outliers. Evaluations were performed using the same protocol as in Fig.~2b, with results summarized over multiple operational states (charging, discharging, and idle) and four LLM backbones (Qwen3‑14B, Llama3.1‑8B, DeepSeek‑v3.2, and ChatGLM‑4.5). Horizontal lines indicate mean FactScore; error bars represent 95\% confidence intervals. ns: not significant; *: $p < 0.1$; **: $p < 0.01$; ***: $p < 0.001$; ****: $p < 0.0001$ (one-sided paired Wilcoxon test).
}\label{si_zju_remain}
\end{figure}

\newpage
\begin{figure}[h!]
\centering
\includegraphics[width=\textwidth]{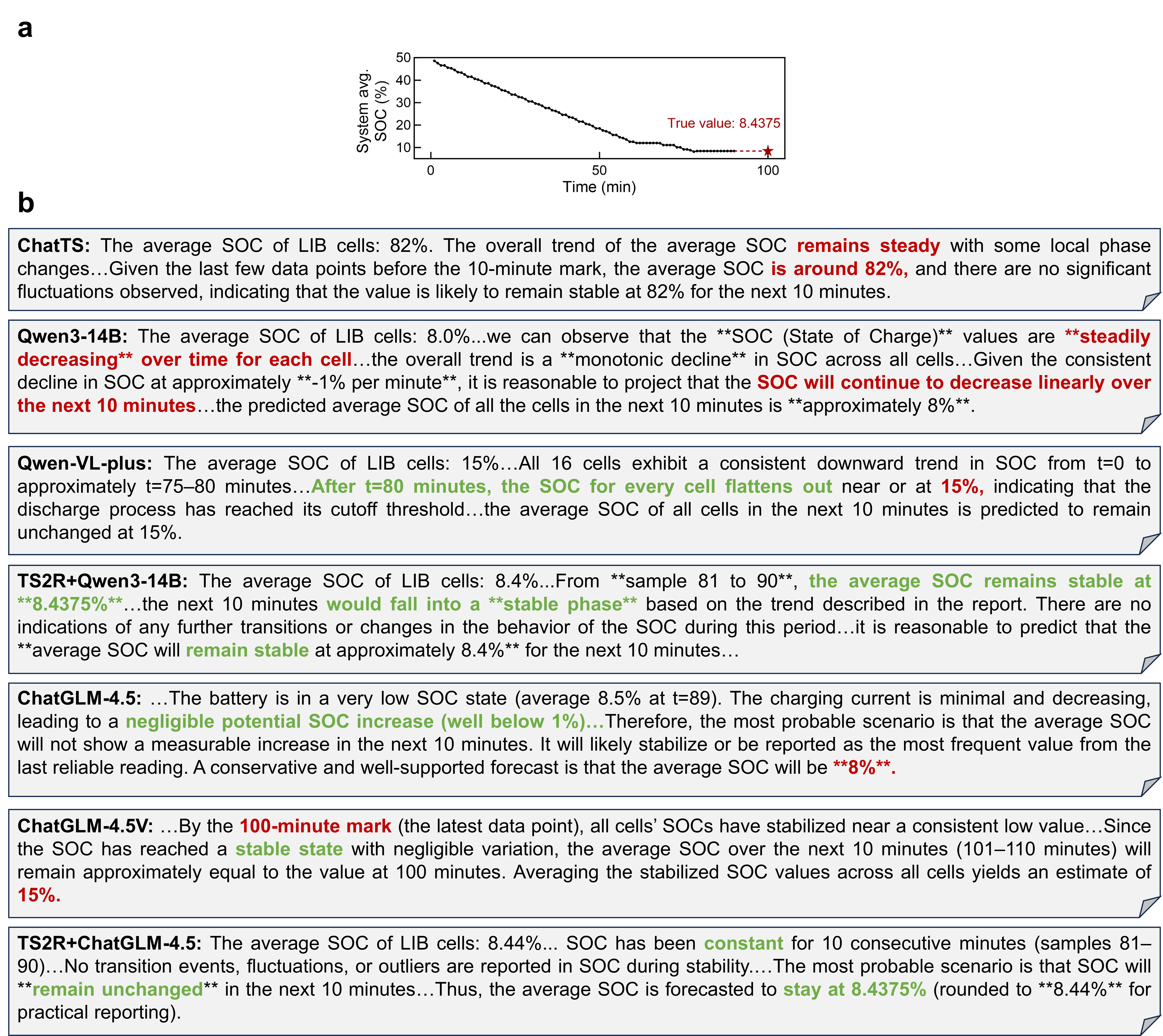}
\caption{Example reports generated by seven LLM-based models (b) for predicting the SOC at minute 100 using data up to minute 90 (panel a). Text highlighted in red indicates statements with low consistency relative to the ground truth, while text in green indicates statements with high consistency.}\label{si_soc}
\end{figure}

\newpage
\begin{figure}[h!]
\centering
\includegraphics[width=\textwidth]{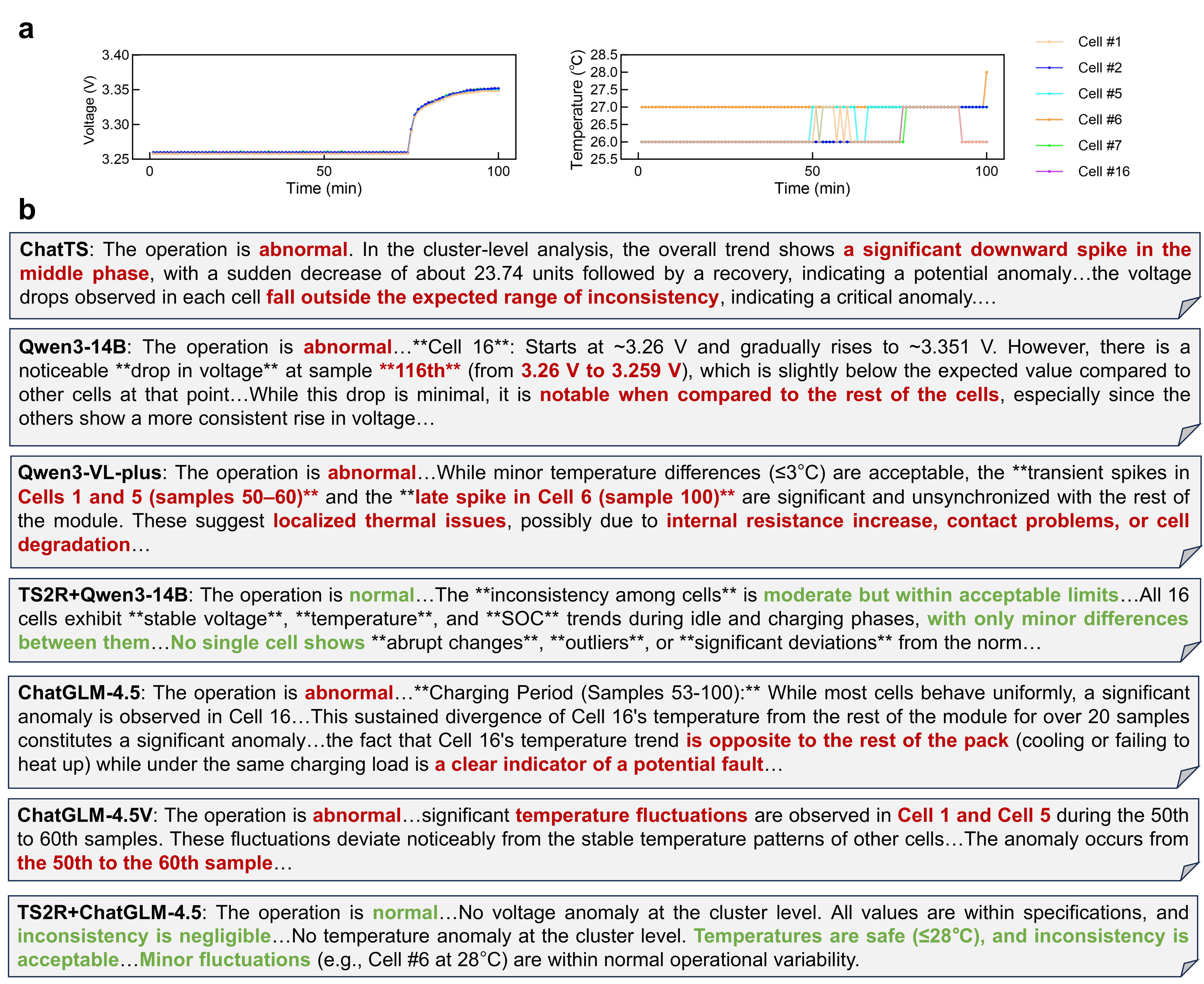}
\caption{Example reports generated by seven LLM-based models (b) for the anomaly-detection task applied to a normally operating sample, with the corresponding voltage and temperature traces of individual LIB cells shown in panel (a). Text highlighted in red indicates statements with low consistency relative to the ground truth, while text in green indicates statements with high consistency.}\label{si_normal}
\end{figure}

\newpage
\begin{figure}[h!]
\centering
\includegraphics[width=\textwidth]{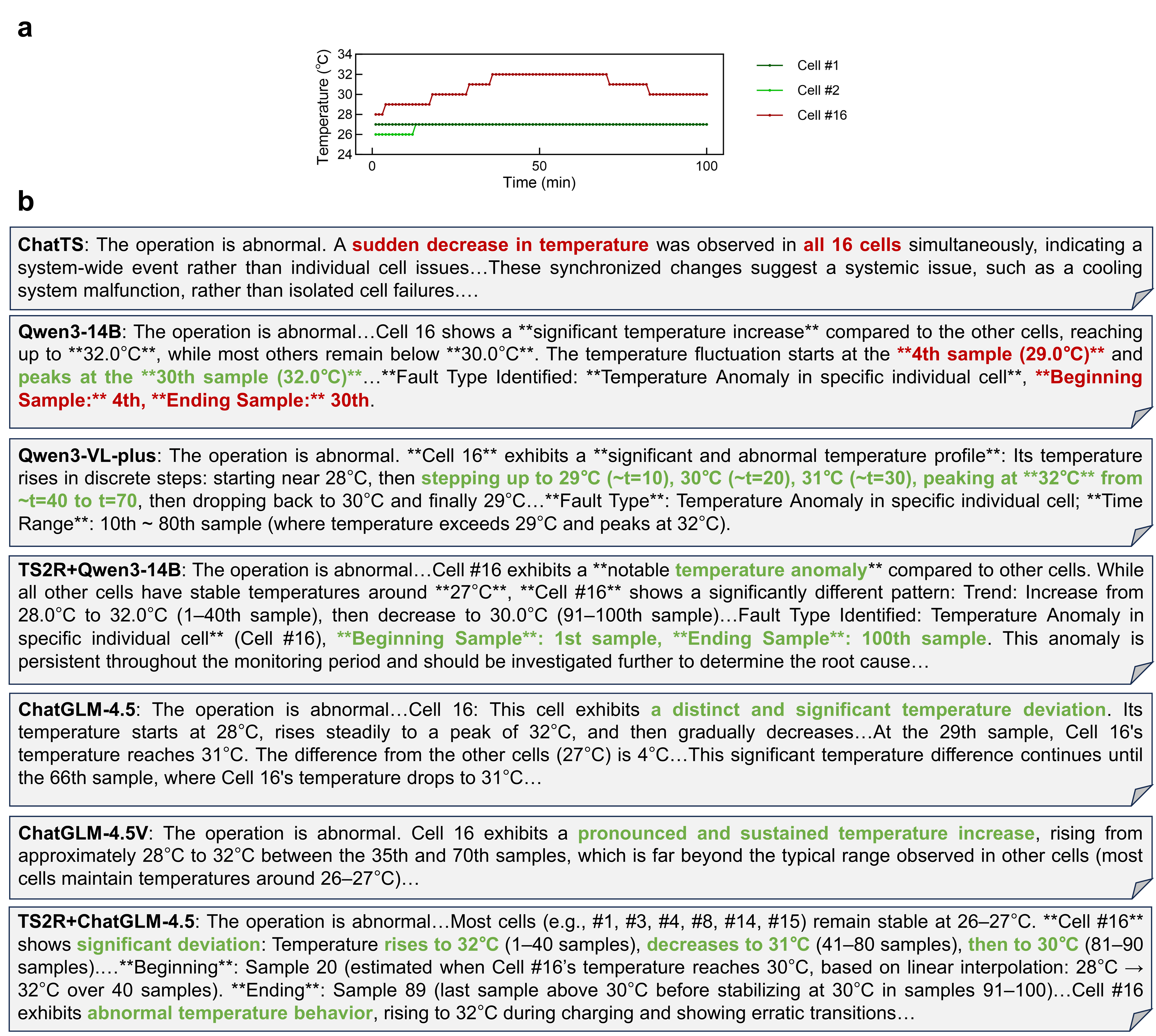}
\caption{Example reports generated by seven LLM-based models (b) for the anomaly-detection task applied to an abnormally operating sample, with the corresponding temperature traces of individual LIB cells shown in panel (a). Text highlighted in red indicates statements with low consistency relative to the ground truth, whereas text in green indicates statements with high consistency.}\label{si_anomaly}
\end{figure}

\newpage
\begin{figure}[h!]
\centering
\includegraphics[width=\textwidth]{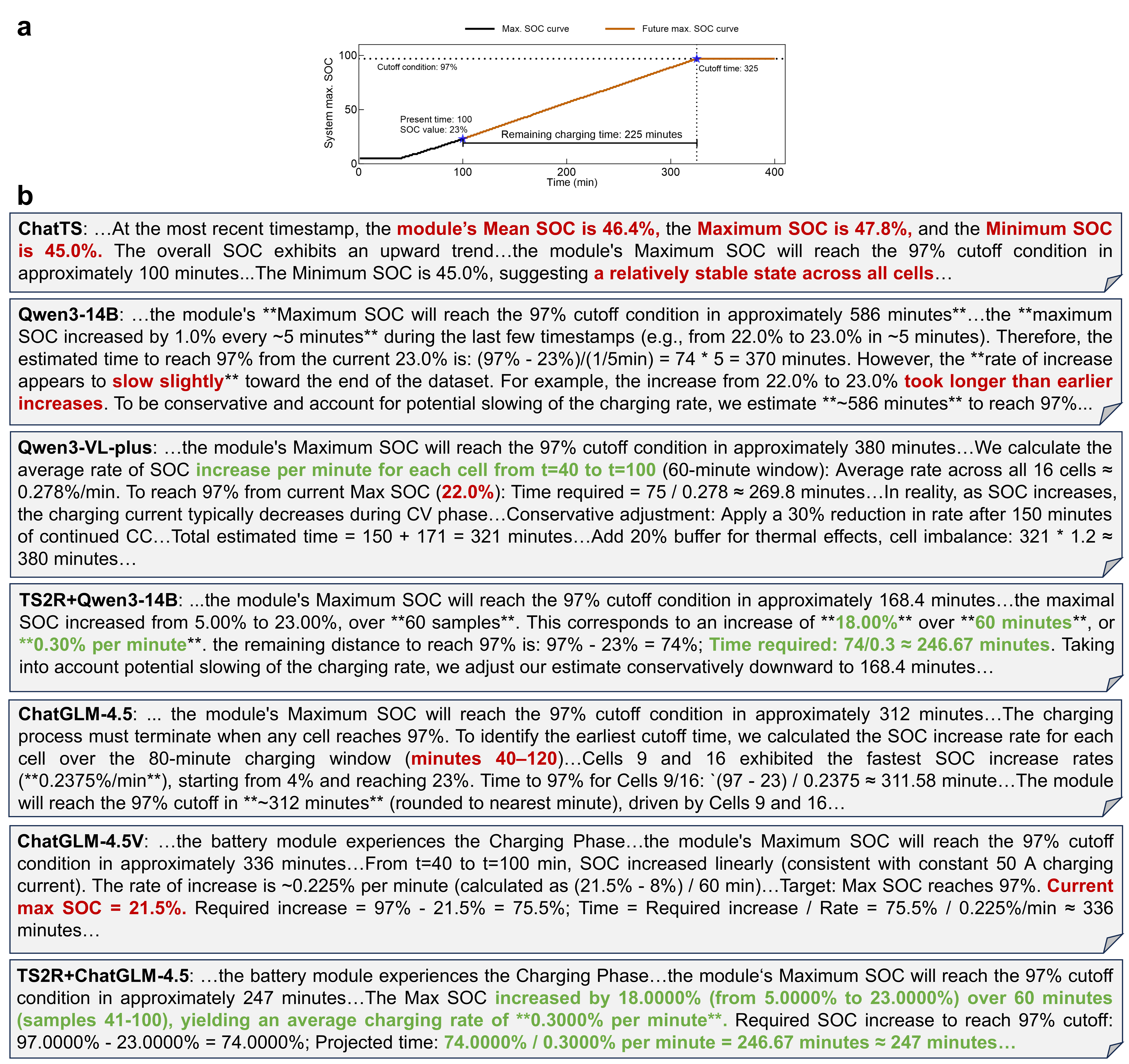}
\caption{Example reports generated by seven LLM-based models (b) for the remaining-charging-time prediction task defined in panel (a). Text highlighted in red denotes statements with low consistency relative to the ground truth, whereas text in green denotes statements with high consistency.
}\label{si_management}
\end{figure}

\newpage
\begin{table}[h]
\centering
\caption{Summary of the three anomaly cases included in the ZJU-ta dataset.}\label{si table: anomaly detection}
\label{tab-s2a} 

\setlength{\tabcolsep}{6pt}
\renewcommand{\arraystretch}{1.25}
\small
\begin{tabularx}{\textwidth}{c| *{4}{>{\centering\arraybackslash}X}}
\hline
\textbf{Case ID} & \textbf{Time period} & \textbf{Module ID} & \textbf{Anomaly type} & \textbf{Description} \\
\hline
1 & 2024-01-22 03:20$\textasciitilde$05:03 & 2 & 
\makecell{Temperature anomaly\\in Cell $\#$16} & 
\makecell{The temperature of \\Cell $\#$16 continuously \\appeared higher than \\others.} \\
\hline
2 & 2024-06-25 13:15$\textasciitilde$15:21 & 1 & 
\makecell{Voltage anomaly\\in Cell $\#$13} & 
\makecell{The voltage of Cell \\$\#$13 shown gradually\\ deviations from \\normal range.} \\
\hline
3 & 2023-09-19 06:27$\textasciitilde$08:10 & 4 & 
\makecell{Voltage anomaly\\in whole system} & 
\makecell{The voltage of cells in\\ Module $\#$4 suddenly\\ dropped.} \\
\hline
\end{tabularx}
\end{table}

\vspace{1cm}

\newpage
\begin{table}[h]
\centering
\centering
\caption{Comparison of the seven methods from Fig.~5 in the main te on three anomaly cases from the ZJU-ta dataset. Text highlighted in red indicates outputs that are inconsistent with the ground truth.} 
\label{tab-s2b} 
\setlength{\tabcolsep}{12pt}
\renewcommand{\arraystretch}{1.8}
\begin{adjustbox}{max width=\textwidth,center}
\small
\begin{tabularx}{\textwidth}{c| *{3}{>{\centering\arraybackslash}X}}
\hline
\textbf{Methods} & \textbf{Abnormal case 1} & \textbf{Abnormal case 2} & \textbf{Abnormal case 3} \\
\hline
\makecell{Ground truth \\of anomaly type}
& \makecell{Temperature anomaly \\in Cell \#16}
& \makecell{Voltage anomaly \\in Cell \#13}
& \makecell{Voltage anomaly \\in whole system} \\
\hline
\makecell{ChatTS}
& \makecell{\color{red}{\textbf{Temperature anomaly}} \\\color{red}{\textbf{in whole system}}}
& \makecell{\color{red}{\textbf{Voltage anomaly}} \\\color{red}{\textbf{in whole system}}}
& \makecell{\color{red}{\textbf{Voltage anomaly}} \\\color{red}{\textbf{in Cell \#1}}} \\
\hline
\makecell{Qwen3-14B}
& \makecell{Temperature anomaly \\in Cell \#16}
& \makecell{Voltage anomaly \\in Cell \#13}
& \makecell{\color{red}{\textbf{Voltage anomaly}} \\\color{red}{\textbf{in Cell \#1}}} \\
\hline
\makecell{Qwen3-VL-Plus}
& \makecell{Temperature anomaly \\in Cell \#16}
& \makecell{Voltage anomaly \\in Cell \#13}
& \makecell{\color{red}{\textbf{Temperature anomaly}} \\\color{red}{\textbf{in Cell \#16}}} \\
\hline
\makecell{TS2R+Qwen3-14B \\ (ours)}
& \makecell{Temperature anomaly\\ in Cell \#16}
& \makecell{Voltage anomaly\\ in Cell \#13}
& \makecell{\color{red}{\textbf{Temperature anomaly}}\\ \color{red}{\textbf{in Cell \#16}}} \\
\hline
\makecell{ChatGLM-4.5}
& \makecell{Temperature anomaly \\ in Cell \#16}
& \makecell{Voltage anomaly \\ in Cell \#13}
& \makecell{Voltage anomaly \\ in whole system} \\
\hline
\makecell{ChatGLM-4.5V}
& \makecell{Temperature anomaly \\in Cell \#16}
& \makecell{Voltage anomaly \\in Cell \#13}
& \makecell{\color{red}{\textbf{Temperature anomaly}} \\\color{red}{\textbf{in Cell \#16}}} \\
\hline
\makecell{TS2R+ChatGLM-4.5 \\ (ours)}
& \makecell{Temperature anomaly\\ in Cell \#16}
& \makecell{Voltage anomaly\\ in Cell \#13}
& \makecell{Voltage anomaly\\ in whole system} \\
\hline
\end{tabularx}
\end{adjustbox}
\end{table}

\newpage
\begin{table}[htbp]
\centering
\caption{Phrasing templates for translating quantitative temporal descriptors into structured textual expressions.}
\label{si_table: phrasing_template}
\vspace*{1em}

\renewcommand{\arraystretch}{1.4}
\setlength{\extrarowheight}{2pt}
\setlength{\tabcolsep}{7pt}

\begin{adjustbox}{max width=\textwidth}
\begin{tabular}{| 
    >{\centering\arraybackslash}m{2.8cm} 
    | >{\raggedright\arraybackslash}m{3.5cm} 
    | >{\raggedright\arraybackslash}m{5cm}
    | >{\raggedright\arraybackslash}m{3.5cm}
    |}
\hline
\bfseries Attributes & \bfseries \makecell{Descriptors} & \bfseries \makecell{Phrasing templates} & \bfseries \makecell{Examples} \\
\hline

\multirow{2}{=}[-25pt]{\centering Trend} & 
\makecell{Increasing;\\Decreasing} & 
From \{begin time\}th to \{end time\}th time, \{time series name\} \{descriptor\} from \{initial value\} to \{final value\}. & 
From 1st to 30th time, voltage of Cell \#1 increases from 3.1V to 3.5V. \\
\cline{2-4}
& 
\makecell{Stable} & 
From \{begin time\}th to \{end time\}th time, \{time series name\} is stable around \{temporal min\}~\{temporal max\} with mean of \{temporal mean\} and std. of \{temporal std\}. & 
From 1st to 30th time, voltage of Cell \#1 is stable around 3.1V~3.4V with mean of 3.1240V and std. of 0.0089. \\
\hline

\multirow{2}{*}{\centering Transition} & 
\makecell{With transition} & 
At \{transition time\}th time, transition points are observed from \{time series name\}. & 
At the 50th and 70th time, transition points are observed from voltage of Cell \#1. \\
\cline{2-4}
& 
\makecell{Without transition} & 
(No translation applied.) & 
\makecell{--} \\
\hline

\multirow{2}{*}{\centering Fluctuation} & 
\makecell{With fluctuation} & 
From \{begin time\}th to \{end time\}th time, \{time series name\} shows fluctuation. & 
From 1st to 5th time, voltage of Cell \#1 shows fluctuation. \\
\cline{2-4}
& 
\makecell{No fluctuation} & 
(No translation applied.) & 
\makecell{--} \\
\hline

\multirow{2}{*}{\centering Outliers} & 
\makecell{With outliers} & 
At time \{occurred time\}, outliers \{outlier values\} are observed from \{time series name\}. & 
At time 11, 14, and 17, outliers (3.7V, 3.75V, 3.7V) are observed from voltage of Cell \#1. \\
\cline{2-4}
& 
\makecell{Without outliers} & 
(No translation applied.) & 
\makecell{--} \\
\hline

\centering Amplitude level & 
\makecell{Significant;\\Moderate;\\Slight} & 
From \{begin time\}th to \{end time\}th time, \{time series name\} shows \{level\} with mean of \{temporal mean\} and std. of \{temporal std\}. & 
From 1st to 50th time, Shannon entropy shows significant level with mean of 1.5550 and std. of 0.0023. \\
\hline

\centering Temporal mean & 
\makecell{--} & 
(Included in expressions for \textit{trend} and \textit{amplitude level}.) & 
\makecell{--} \\
\hline
\centering Temporal std & 
\makecell{--} & 
(Included in expressions for \textit{trend} and \textit{amplitude level}.) & 
\makecell{--} \\
\hline
\centering Initial value & 
\makecell{--} & 
(Included in expressions for \textit{trend} attribute.) & 
\makecell{--} \\
\hline
\centering Final value & 
\makecell{--} & 
(Included in expressions for \textit{trend} attribute.) & 
\makecell{--} \\
\hline
\end{tabular}
\end{adjustbox}
\end{table}